\documentclass[11pt]{article}

\usepackage[final]{acl}

\usepackage{times}
\usepackage{latexsym}
\usepackage[T1]{fontenc}

\usepackage[utf8]{inputenc}

\usepackage{microtype}

\usepackage{inconsolata}

\usepackage{graphicx}

\usepackage[normalem]{ulem}
\usepackage{paralist}
\usepackage{booktabs}
\usepackage{multirow}
\usepackage[most]{tcolorbox}
\usepackage{fvextra}
\usepackage{amsmath}
\usepackage{pifont}
\usepackage{makecell}
\usepackage{hyperref}
\usepackage{url}
\title{From Memory to Behavior: \\A Behavior-Aware Role-Playing Framework for Social Media Influencers}

\author{
Ji-Lun Peng$^{\star \dag}$\quad 
Yi-Zhen Zhang$^\dag$\quad
Chun-Nan Chou$^\dag$\quad
Yun-Nung Chen$^\star$
\\
 $^\star$National Taiwan University, Taipei, Taiwan \\
 $^\dag$CMoney Technology Corporation, Taiwan
\\
   \texttt{r13946006@ntu.edu.tw}\quad \texttt{\{jason\_zhang,jason\_chou\}@cmoney.com.tw} \\ 
   \texttt{y.v.chen@ieee.org}
}
\begin{document}
\maketitle
\begin{abstract}
Large language models have shown strong potential as role-playing agents for real individuals, yet faithful impersonating remains challenging. Existing in-context learning-based methods fail to capture how individuals react under different situations. In addition, LLM-based evaluation is difficult for obscure individuals. To address these challenges, we propose \textbf{Situation--Internal state--Behavior Persona} method to incorporate situation-dependent behavioral strategies. We further design an evaluation protocol that provides LLM evaluators with references about the impersonated individual. We evaluate our approach on a newly constructed dataset for the task of generating replies on social media. Experimental results show that our proposed method outperforms state-of-the-art ICL-based baselines, while our evaluation protocol achieves moderate correlation with human judgment.
Besides, experiments on fictional-character benchmarks demonstrate that our proposed method is applicable beyond the social media setting. 
These findings suggest that incorporating behavioral information broadly improves the fidelity of role-playing for real individuals on social media or fictional characters.\footnote{Code: \url{https://github.com/MiuLab/SIBPersona}}

\end{abstract}
\section{Introduction}
Recent advances in large language models (LLMs) have accelerated the development of role-playing agents (RPAs) capable of impersonating both fictional characters \citep{yu2025beyond, zhang2025omnicharacter} and real individuals \citep{chuang2025debate, wang2025survey}. 
Recent evidence further suggests that RPAs for real individuals can produce convincing impersonations \citep{shi2025impersona}.

\begin{figure}[t!]
\centering
    \includegraphics[width=0.48\textwidth]{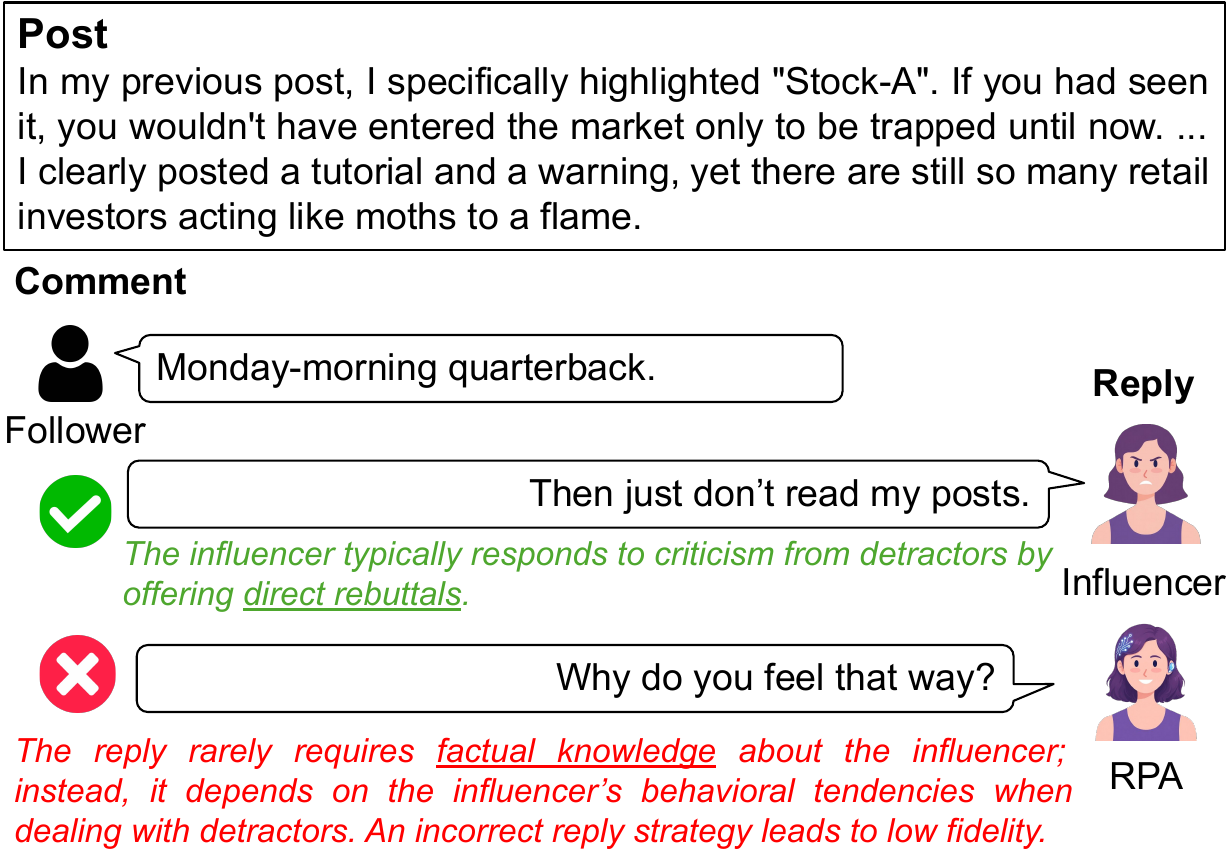}
    \caption{In the social media comment-replying scenarios between influencers and followers, replies involve short texts where behavioral strategy is more important for determining the fidelity of impersonation.}
    \label{figure:example_data}
\end{figure}

Toward faithful role-playing of specific real individuals, two major challenges remain. 
\textbf{Behavioral Modeling Gap:} Existing in-context learning (ICL)-based role-playing methods mainly provide RPAs with factual information about impersonated individuals extracted from historical conversations \citep{liu2024roleagent, shi2025impersona, gao2025tailorrpa}. 
However, such information does not explicitly capture how an individual tends to react under different situations. 
As illustrated in \autoref{figure:example_data}, in social media comment-replying scenarios between influencers and followers, replies are often brief or knowledge-light, making faithful impersonation depends more on behavioral strategies than on factual information.
\textbf{LLM Familiarity Gap:} Existing LLM-based evaluation protocols for role-playing are largely developed for fictional characters, whose information are often represented in LLMs' knowledge \citep{peng2026rethinking}. 
As a result, directly prompting an LLM can provide a reasonable evaluation \citep{wang2024rolellm, yu2025beyond}. 
By contrast, this direct prompting strategy is less applicable to obscure real individuals, whose information may be sparsely represented in LLMs' pre-training data. 
This limited familiarity makes it difficult to directly use LLMs for automated evaluation of role-playing real-individual. 

To address these challenges, we develop a behavior-aware framework for role-playing real individuals. 
Inspired by the Cognitive-Affective Processing System \citep{mischel1995cognitive}, we propose \textbf{Situation–Internal state–Behavior Persona (SIBPersona)}, a method that models individuals' situation-dependent behavioral strategies.
In addition, we design an automated LLM-based evaluation protocol tailored to role-playing real-individual, which provides the LLM evaluator with reference about the impersonated individual to alleviate its limited familiarity with the individual.
We apply this framework to the domain of social media by constructing a dataset of influencer–follower interactions.
Using this dataset, we evaluate RPAs on their ability to impersonate influencers, specifically in generating replies to followers' comments.

The contributions of this paper are three-fold:
\begin{compactitem}
    \item We present a structured view of persona for RPAs by distinguishing data-invariant and data-variant personae. Building on this view, we propose SIBPersona, which introduces reaction process as a novel dimension of data-variant persona to capture situation-dependent behavioral strategies.
    \item We establish an reference-augmented evaluation protocol for role-playing of obscure individuals. The protocol achieves moderate correlation with human judgment.
    \item We conduct comprehensive experiments across multiple influencers, model families, and fictional-character benchmarks, showing that SIBPersona consistently outperforms state-of-the-art ICL-based baselines.
\end{compactitem}

\section{Related Work}
\paragraph{Approaches of LLM Role-Playing}
Two mainstream approaches enable LLMs to impersonate specific targets: Fine-Tuning (FT) \citep{yang2025hycora, ye2025cpo} and ICL.
We focus on ICL-based methods because persona information is explicitly represented in prompts, making the role-playing process easier to interpret and modify. Existing ICL-based methods utilize retrieval-augmented generation (RAG) \citep{lewis2020retrieval} to incorporate context-relevant information into prompts \citep{shi2025impersona, gao2025tailorrpa}. However, the retrieved content primarily consists of raw conversational data or factual knowledge. Such approaches largely emphasize what an individual knows rather than how an individual tends to behave across different interaction situations.

To address this gap, we draw on the Cognitive-Affective Processing System (CAPS) \citep{mischel1995cognitive}, which models behavior as a dynamic reaction process. According to CAPS, individuals perform behaviors by processing situational inputs through internal states. This perspective suggests that RPAs should model the situation-dependent reaction process. Nonetheless, how to explicitly model such a dynamic reaction process in ICL-based method remains underexplored.

\paragraph{Datasets for Role-Playing of Real Individuals}
Existing role-playing datasets primarily focus on fictional characters \citep{xu2024character, xu2026adamarp}, while datasets for role-playing real individuals remain relatively limited. 
To study role-playing real-individual, we turn to social media, which provides rich conversational records for observing how an individual responds across diverse situations \citep{li2025personalized}.
To support our research goal of impersonating influencers, the dataset must satisfy two requirements: (1) it should contain abundant influencer-centered interactions from a single identifiable individual, allowing LLMs to obtain comprehensive information about the impersonated individual; and (2) it should preserve the comment-reply structure under the influencer's posts, which is essential for observing how the individual responds to different situations. 

As summarized in \autoref{table:dataset_comparison}, no existing dataset satisfies both requirements simultaneously. 
Among the public datasets, BluePrint \citep{buck2025texttt} SYNTHIA \citep{rahimzadeh2025synthia}, PersonalDialog \citep{zheng2019personalized}, and Persona-Chat \citep{zhang2018personalizing} do not preserve an identifiable individual's history: BluePrint anonymizes users and aggregates interactions into clusters, while SYNTHIA provides only synthesized persona attributes rather than raw, individual-level social media interactions. 
PER-CHAT \citep{wu2021personalized} does contain large-scale, identifiable individual histories ($314,749$ individuals), but lacks the comment-reply structure needed to observe situational responses.
Conversely, PersonaConvBench \citep{li2025personalized} preserves this structure but provides a limited number of samples per individual ($6.87$ posts and $37.05$ comment-reply pairs on average), making it insufficient for LLMs to obtain comprehensive information about the impersonated individual.
These limitations motivate us to construct a new dataset tailored for the task of generating replies on social media.

\begin{table*}[hbt!]
    \centering
    \setlength{\tabcolsep}{3pt}
    \resizebox{\linewidth}{!}{%
    \begin{tabular}{l|cccccccc}
    \toprule
Dataset & Public & \makecell{Identifiable\\Individual\\ History} & \#Individual & \makecell{Comment-\\Reply\\ Structure}  
& \makecell{Avg. Posts\\per Individual} 
& \makecell{Avg.\\ Comment-Reply\\Pairs per\\ Individual} 
& \makecell{Max Posts\\per Individual} 
& \makecell{Max\\ Comment-Reply\\Pairs per\\ Individual} \\
      \midrule
BuzzFace \shortcite{santia2018buzzface} & \ding{55} & - & - & - & - & - & - & - \\
TWON \shortcite{munker2026don} & \ding{55} & - & - & - & - & - & - & - \\
BluePrint \shortcite{buck2025texttt} & \ding{51} & \ding{55} & - & - & - & - & - & - \\
SYNTHIA \shortcite{rahimzadeh2025synthia} & \ding{51} & \ding{55} & - & - & - & - & - & -\\
PersonalDialog \shortcite{zheng2019personalized} & \ding{51} & \ding{55} & - & - & - & - & - & -\\
Persona-Chat \shortcite{zhang2018personalizing}  & \ding{51} & \ding{55} & - & - & - & - & - & -\\
PER-CHAT \shortcite{wu2021personalized}  & \ding{51} & \ding{51} & 314749 & \ding{55} & - & - & - & -\\
PersonaConvBench \shortcite{li2025personalized}  & \ding{51} & \ding{51} & 2796 & \ding{51}  & 6.87 & 37.05 & 111 & 414\\
\midrule
\bf Ours & \ding{51} & \ding{51} & 6  & \ding{51}  &\bf 1811 &\bf 5554 &\bf 2903 &\bf 15240\\
\bottomrule
    \end{tabular}
    }
    \caption{Feasibility of existing datasets for studying role-playing of real individuals on social media reply generation. ''-'' indicates that the corresponding information is not applicable under our comparison criteria. Since our research goal is to impersonate influencers across diverse interaction situations, we construct a dataset centered on a small number of influencers with rich historical interactions.}
    \label{table:dataset_comparison}
\end{table*}

\paragraph{Evaluation of RPAs}
\label{sec:eval}
LLM evaluation have become widely adopted for evaluating role-playing of fictional character \citep{gu2024survey}, as they enable fine-grained assessment of performance \citep{wang2025characterbox, wang2024rolellm, shao2023character}. These approaches typically assume that the evaluator model possesses sufficient knowledge about the impersonated character \cite{peng2026rethinking}.

Evaluating role-playing of real individuals remains challenging because LLM evaluators have limited knowledge about obscure people. Consequently, prior studies on role-playing real individuals often rely on human evaluation \citep{shi2025impersona, tu2023characterchat}, while automated evaluation protocols for role-playing of obscure real individuals remain underdeveloped.

\section{SIBPersona}
In this section, we present \textbf{SIBPersona} method designed for role-playing of influencers on social media.
We first introduce our multifaceted view of persona and define the reaction process in \S\ref{sec:nature_of_persona}, followed by formalizing the task of generating replies on social media in \S\ref{sec:problem_definition}.
Subsequently, we elaborate how data-variant persona is constructed from historical data in \S\ref{sec:persona_corpus}.
Finally, we explain how the constructed data-variant persona is retrieved and used for reply generation in \S\ref{sec:reply_generation}.

\subsection{Multifaceted Nature of Persona}
\label{sec:nature_of_persona}
In RPA, a "persona" is always referred to the information describing fictional characters or real individuals.
We argue that personae can be divided into two types: \textbf{data-invariant} and \textbf{data-variant}.
The data-invariant persona, as the name implies, is provided in advance and is not derived from the data. 
For example, we can create a narrative to describe the demographic information of the influencer or the fictional traits of the fictional character, as far as we know.
On the other hand, the data-variant persona is derived from the data and designated to capture the dynamics of RPA.

Apart from deriving from the data or not, the prior methods mainly modeled the persona by two dimensions: 1) memory ~\cite{shi2025impersona} capturing the factual information and 2) speaking style ~\cite{tu2024charactereval} capturing expression habits.
However, we contend that these two dimensions are insufficient for RPAs.
As suggested in CAPS, responses of real people are shaped not only by what they know and how they speak but also by the reaction process they tend to follow in a given situation.

\paragraph{Reaction Process}
\label{sec:reaction_process}
In order to resolve this issue, we propose to incorporate so-called reaction process, the third dimension, into the data-variant persona.
To this end, we model reaction process as the mapping of a given situation to an internal state and subsequently to a behavioral strategy. 
Formally, reaction process is defined by $\{\langle s,i,b \rangle \in S \times I \times B \mid s = LLM(x), i = LLM(x), b = LLM(x) \}$, where $S$ contains situations, $I$ represents the set of internal states, $B$ is the set of behavioral strategies, and $x$ denotes a text input.
We call this three-tuple, namely $\langle s,i,b \rangle$, an SIB triplet.
In the following, we utilize the example that influencers reply followers' comments on social media to illustrate the idea of reaction process.
When influencers reply to followers' comments, they first interpret the comments into high-level situations, e.g., praise, inquiry, or hostility. 
This emerging situation then triggers a specific internal state, such as happy or angry, which leads to a certain behavioral strategy, such as expressing gratitude or defensive rebuffing. 
Grounded in CAPS, this three-dimensional data-variant persona provides a more psychologically faithful modeling of how real individuals respond.

\begin{figure*}[hbt!]
\centering
    \includegraphics[width=0.95\textwidth]{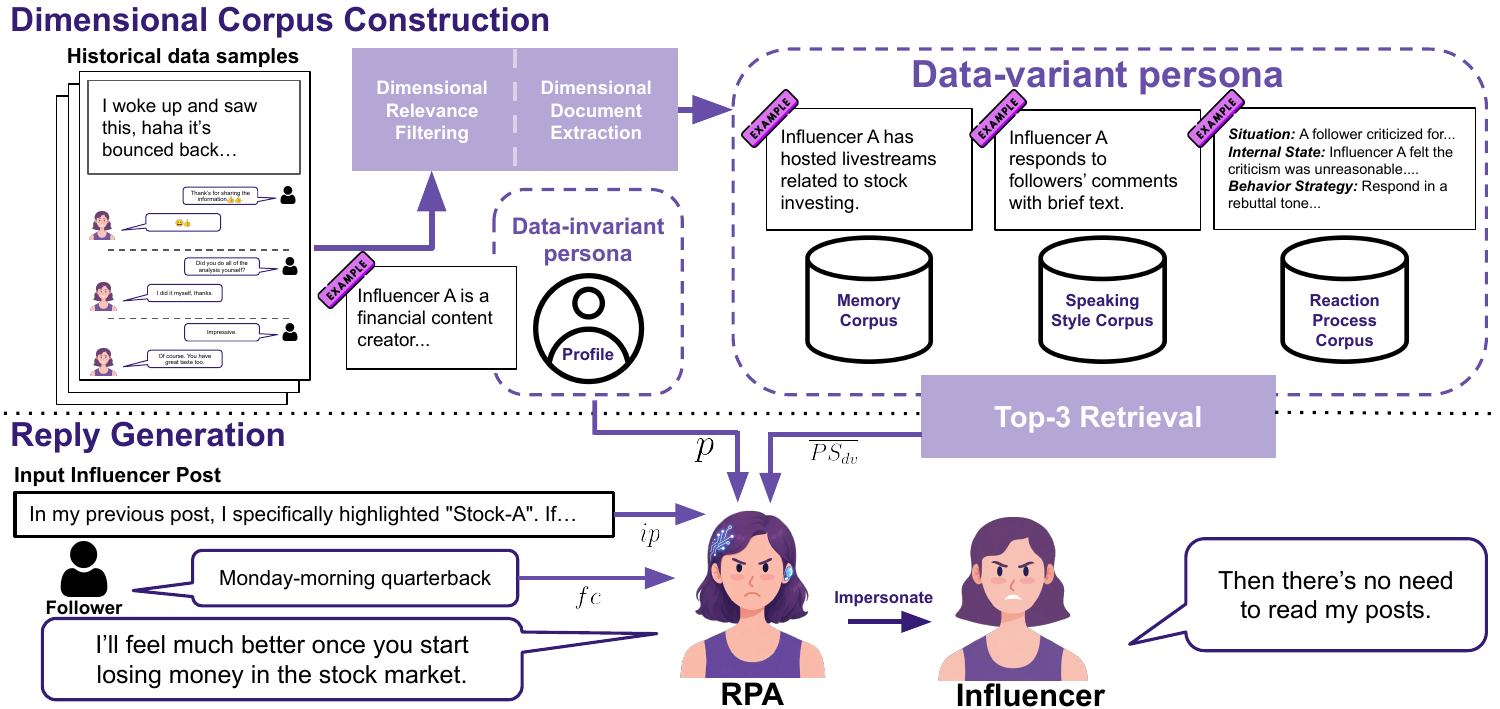}
    \caption{Overview of the SIBPersona method. The upper part illustrates the process of extracting data-variant persona from historical data. The lower part presents reply generation process: retrieving the most relevant dimensional documents based on the comment enables the RPA to generate a high-fidelity reply.}
    \label{figure:framework}
\end{figure*}

\subsection{Problem Formulation}
\label{sec:problem_definition}
After illustrating the three-dimensional data-variant persona, we provide the formal settings of the problem for easy exposition.
In general, RPAs are designed to engage in conversations with users by emulating specific characters or impersonating specific individuals.
In this work, we focus on the generation of single-turn replies for influencers on social media.
To be specific, we generate the reply $r$ according to the influencer's post $ip$, the follower's comment $fc$, the influencer's data-invariant profile $p$, and the influencer's data-variant persona $PS_{dv}$, which can be represented as:
$$r=RPA(ip, fc, p, PS_{dv}).$$
In particular, $PS_{dv}$ contains information about the three aforementioned dimensions.

\subsection{Constructing Data-Variant Persona}
\label{sec:persona_corpus}
In this subsection, we elaborate on how each dimension is extracted and represented.
We first view a data sample associated with a specific influencer as $ds_j = \langle ip_j, CRP_j \rangle$,
where $ip_j$ is the influencer's post, and $CRP_j = \{ \langle fc_{j1}, r_{j1} \rangle, \langle fc_{j2}, r_{j2} \rangle, \dots, \langle fc_{jn}, r_{jn} \rangle \}$ denotes the set of all comment-reply pairs associated with $ip_j$ post.
As a consequence, the historical data of this influencer is a collection of the data samples, denoted as $\mathcal{D} = \{ds_1, ds_2, \dots, ds_n\}$.

In accordance with the previous works of extracting data-variant personae by LLMs \citep{shi2025impersona, liu2024roleagent}, we employ a two-stage pipeline to produce data-variant personae from raw social media histories.
Our entire pipeline is depicted in the upper part of \autoref{figure:framework} and is explained in detail below.

\paragraph{Stage 1: Dimensional Relevance Filtering}
Due to data noise on social media, we employ a recognizing LLM to filter irrelevant information in the first stage to increase the efficacy of the data pipeline.
Specifically, the recognizing LLM acts like a multi-label classifier that assigns each data sample to the set of dimensional labels $L=\{l_M, l_T, l_{RP}\}$. 
That is, $ \forall j \ h(ds_j)=\mathcal{A}_{j} \ \ \mathcal{A}_{j} \subseteq L$, where $h$ is a mapping function from a data sample to the subset of $L$.
As a shorthand, we use the notations $M$, $T$, $RP$ to represent memory, speaking style, and reaction process, respectively.
If the recognizing LLM assigns the data sample $ds_j$ to the empty set, i.e. $h(ds_j)=\emptyset$, $ds_j$ is discarded.
Otherwise, $ds_j$ proceeds to the next stage.

\paragraph{Stage 2: Dimensional Document Extraction}
Given a data sample $ds_j = \langle ip_j, CRP_j \rangle$ such that $h(ds_j)\neq \emptyset$, we use an extracting LLM to generate the dimensional information of $ds_j$ for each label $l \in h(ds_j)$.
We introduce \textbf{dimensional documents} whose details are elaborated later as a countable surrogate for the generated dimensional information.
Given $ds_j$ and a label $l \in h(ds_j)$, the extracting LLM may generate more than one dimensional document, and we use $k$ to index the generated dimensional documents.

To equip our approach with interpretability, we define a subset of $CRP_j$ as a set of \textbf{evidence pairs} denoted $EP_{jk}$, and $EP_{jk}$ provides the grounding evidence for the generated dimensional information.
If the label $l$ is $l_M$, the extracting LLM generates a memory document $d_{jk}^{(l_M)} = \langle mp_{jk}, EP_{jk} \rangle$, where $mp_{jk}$ denotes a memory point capturing factual information.
When the label $l$ is $l_T$, the extracting LLM produces a speaking style document $d_{jk}^{(l_T)} = \langle sp_{jk}, EP_{jk} \rangle$, where $sp_{jk}$ represents a speaking style point reflecting a expression habit.
Otherwise, the extracting LLM yields a reaction process document $d_{jk}^{(l_{RP})} = \langle \langle s_{jk}, i_{jk}, b_{jk} \rangle, EP_{jk} \rangle$, where $\langle s_{jk}, i_{jk}, b_{jk} \rangle$ constitutes an SIB triplet comprising the situation, internal state, and behavioral strategy, respectively.

Depending on the type of dimensions, we store the extracted dimensional document in the corresponding \textbf{dimensional corpus}, which is a collection of dimensional documents.
Thus, our data-variant persona $PS_{dv}$ has three dimensional corpora: memory corpus, speaking style corpus, and reaction process corpus.
These three dimensional corpora lay the foundation for the reply generation illustrated in the next subsection.

\subsection{Reply Generation}
\label{sec:reply_generation}
After the data-variant persona is established, one intuitive idea is to generate the reply by considering all dimensional documents in these three dimensional corpora as the demonstrations to LLMs.
However, this idea is inherently infeasible since influencers on social media always have numerous posts and comment-reply pairs; otherwise, they could hardly be called "influencers."
Instead of using $PS_{dv}$ directly, we utilize the approximate data-variant persona $\overline{{PS_{dv}}}$ derived from $PS_{dv}$ as an alternative.
 
As depicted in the lower part of \autoref{figure:framework}, we utilize the concept of RAG \citep{lewis2020retrieval} to derive $\overline{{PS_{dv}}}$.
Given the query $fc$, we independently retrieve the top three dimensional documents which are most relevant to the query $fc$ from each dimensional corpus.
The relevance measure is calculated as follows.
We treat $mp_{jk}$, $sp_{jk}$, and $s_{jk}$ in the SIB triplet as the retrieval keys for memory corpus, speaking style corpus, and reaction process corpus, respectively.
For each dimensional corpus, we compute the cosine similarity between the embedding of query $fc$ and the embedding of the retrieval keys.
Afterwards, we rank the dimensional documents by similarity scores and retrieve the top three corresponding dimensional documents for each dimensional corpus.
Finally, the nine retrieved dimensional documents constitute $\overline{{PS_{dv}}}$, serving as demonstrations to LLM. The problem we address becomes: $$r=RPA(ip, fc, p, \overline{PS_{dv}}).$$ \autoref{figure:case_english} shows an example of a retrieved dimensional document and the reply generated by SIBPersona. Additional implementation details are provided in Appendix \ref{appendix:SIBPersona}.

\section{Reference-Augmented Evaluation}
\label{sec:evaluation_protocol}
In this section, we propose a reference-augmented protocol for automatically evaluating role-playing agents. 
We first explain the challenges of this task, then detail our design, and conclude by empirically validating its alignment with human judgment.

\begin{table*}[t!]
    \centering
    \small
    \setlength{\tabcolsep}{2pt}
    \begin{tabular}{l|ccc|ccc|ccc|ccc|ccc}
    \toprule
    \bf Methods 
    & \multicolumn{3}{c|}{\bf GPT-5.2} 
    & \multicolumn{3}{c|}{\bf Gemini-3-flash} 
    & \multicolumn{3}{c|}{\bf DeepSeek-V3.2} 
    & \multicolumn{3}{c|}{\bf Claude-Opus-4.6} 
    & \multicolumn{3}{c}{\bf Average} \\
    
    & $M$ & $T$ & $RP$
    & $M$ & $T$ & $RP$
    & $M$ & $T$ & $RP$
    & $M$ & $T$ & $RP$
    & $M$ & $T$ & $RP$ \\
    \midrule

    Naive
    & \bf 97.50 & 46.40$^\ddag$ & 67.53$^\ddag$
    & 94.70 & 42.47$^\ddag$ & 69.27$^\ddag$
    & 95.13 & 48.93$^\ddag$ & 70.83$^\ddag$
    & 96.77 & 46.93$^\ddag$ & 76.87$^\ddag$
    & 96.03 & 46.18$^\ddag$ & 70.78$^\ddag$ \\
    
    CoT 
    & 97.47 & 44.93$^\ddag$ & 70.73$^\ddag$
    & \bf 95.83 & 41.17$^\ddag$ & 69.53$^\ddag$
    & \bf 96.27 & 49.23$^\ddag$ & 71.47$^\ddag$
    & 97.33 & 47.00 & 77.33$^\ddag$
    & \bf 96.73 & 45.58$^\ddag$ & 72.27$^\ddag$ \\
    
    IMPersona 
    & 96.07 & 47.50$^\ddag$ & 71.93$^\ddag$
    & 94.90 & 44.60$^\ddag$ & 70.10$^\ddag$
    & 96.00 & 50.80$^\ddag$ & 72.50$^\ddag$
    & \bf 97.43 & 48.67$^\ddag$ & 78.13$^\ddag$
    & 96.10 & 47.89$^\ddag$ & 73.17$^\ddag$ \\
    
    Vanilla RAG 
    & 97.33 & 49.43$^\ddag$ & 73.20$^\ddag$
    & 93.77 & 53.20 & 71.50$^\ddag$
    & 94.23 & 58.37$^\ddag$ & 73.43
    & 96.70 & 56.93 & 77.70$^\ddag$
    & 95.51 & 54.48$^\ddag$ & 73.96$^\ddag$ \\
    
    SIBPersona 
    & 96.90 & \bf 54.33 & \bf 77.63
    & 93.77 & \bf 56.47 & \bf 75.50
    & 94.83 & \bf 63.67 & \bf 76.43
    & 96.93 & \bf 62.77 & \bf 81.93
    & 95.61 & \bf 59.31 & \bf 77.88 \\
    
    \bottomrule
    \end{tabular}
    \caption{Performance on \textit{Influencer A}. \textbf{Bold}: the best results; $\ddagger$: significant degradation compared to SIBPersona ($p<0.05$).}
    \label{table:main_model}
\end{table*}

\begin{table*}[t!]
    \centering\small
    \begin{tabular}{l|ccc|ccc}
    \toprule
    \bf Methods
    & \multicolumn{3}{c|}{\bf Deepseek-V3.2}
    & \multicolumn{3}{c}{\bf Gemini-3-flash} \\
    
    &\bf $M$ &\bf $T$ &\bf $RP$
    &\bf $M$ &\bf $T$ &\bf $RP$ \\
    \midrule

    Naive
    & 92.10 $\pm$ 3.04
    & 42.03$^\ddagger$ $\pm$ 8.43
    & 68.48$^\ddagger$ $\pm$ 5.86

    & {\bf 92.50} $\pm$ 2.32
    & 43.33$^\ddagger$ $\pm$ 7.35
    & 70.64$^\ddagger$ $\pm$ 7.37\\
    
    CoT
    & {\bf 92.17} $\pm$ 3.03
    & 43.92$^\ddagger$ $\pm$ 8.13
    & 69.83$^\ddagger$ $\pm$ 6.14

    & 92.15 $\pm$ 2.54
    & 44.05$^\ddagger$ $\pm$ 7.08
    & 70.48$^\ddagger$ $\pm$ 6.40\\
    
    IMPersona
    & 91.60 $\pm$ 2.21
    & 45.54$^\ddagger$ $\pm$ 5.86
    & 71.14$^\ddagger$ $\pm$ 6.02

    & 91.42 $\pm$ 1.94
    & 45.21$^\ddagger$ $\pm$ 6.40
    & 71.56$^\ddagger$ $\pm$ 7.19\\
    
    Vanilla RAG
    & 91.39 $\pm$ 3.23
    & 57.05$^\ddagger$ $\pm$ 2.43
    & 72.43$^\ddagger$ $\pm$ 5.06

    & 92.19 $\pm$ 2.46
    & 53.43$^\ddagger$ $\pm$ 4.92
    & 72.59$^\ddagger$ $\pm$ 4.79\\
    
    SIBPersona
    & 92.14 $\pm$ 3.26
    & {\bf 61.09} $\pm$ 1.66
    & {\bf 75.84} $\pm$ 4.90

    & 92.21 $\pm$ 2.20
    & {\bf 57.34} $\pm$ 3.20
    & {\bf 76.48} $\pm$ 5.08\\
    
    \bottomrule
    \end{tabular}
    \caption{Results on other five influencers. \textbf{Bold} indicates the better scores. $\ddagger$: significant degradation compared to SIBPersona ($p<0.05$).}
    \label{table:main_people}
\end{table*}

\begin{table}[t!]
    \centering
    \small
    \setlength{\tabcolsep}{3pt}
    \begin{tabular}{l|ccc}
    \toprule
   \bf Methods & \bf $M$ & \bf $T$ & \bf $RP$  \\
      \midrule
SIBPersona & 92.67 $\pm$ 3.18 & \bf 61.60 $\pm$ 1.82 & \bf 75.96 $\pm$ 4.49  \\
w/o $I$ & 92.61 $\pm$ 2.49 & 60.30 $\pm$ 2.37 & 75.44 $\pm$ 3.61  \\
w/o $RP$ & \bf 92.83 $\pm$ 2.44 & 57.82$^\ddagger$ $\pm$ 2.51 & 73.36$^\ddagger$ $\pm$ 4.64  \\
\bottomrule
    \end{tabular}
    \caption{Dimension ablation study using \texttt{DeepSeek-V3.2} on the six influencer. \textbf{Bold}: the best results; $\ddagger$: significant degradation compared to SIBPersona ($p<0.05$).}
    \label{table:influencer_ablation}
\end{table}

\subsection{Penalty-Based LLM Evaluator}
To enable scalability and facilitate future iterations, we adopt LLM evaluators to automatically assess role-playing agents for the task of generating replies on social media.
However, as addressed in \S\ref{sec:eval}, LLM evaluators may have limited knowledge of obscure individuals.
To resolve this knowledge scarcity issue, we augment the LLM evaluator with some references called \textbf{representative dimensional documents} about the impersonated individual.
The representative dimensional documents are constructed as follows.
We first apply the method described in \S\ref{sec:persona_corpus} to construct the three dimensional corpora from the test samples and then select representative dimensional documents, denoted as $rdd_l$, from the resulting dimensional corpus for each dimension $l \in L$.
Specifically, we apply K-medoids clustering to select $\sqrt{M/2}$ representative dimensional documents from each dimensional corpus, where $M$ is the size of dimensional documents.
Clustering is performed based on the embeddings of $s_{jk}$, $mp_{jk}$ and $sp_{jk}$.
Crucially, the representative dimensional documents are constructed and sampled exclusively from the test samples rather than the demonstration examples.
This ensures that the LLM evaluator's reference standard remains independent of the role-playing agent's demonstration source, so that the evaluation reflects the agent's true generalization ability rather than its capacity to reproduce seen demonstrations.

For the clarity of presentation, we define a test sample as $ts_j = \langle ip_j, fc_j, r_{j}^{a} \rangle,$ where $r_{j}^{a}$ is the reply generated by an RPA. 
For each dimension $l \in L$, our LLM evaluator takes $ts_j$, $p$, $rdd_l$, and the ground-truth reply $r_{j}^{gt}$ written by the influencer as input. 
Following \citet{wang2025coser}, our LLM evaluator is instructed to identify the flaws in $r_{j}^{a}$ and to assign a severity level ranging from $1$ to $5$ for each flaw at the same time. 
The final score for the dimension $l$ is computed by averaging the scores on all test samples whose size is $N$, which can be summarized as follows:
$$
Score_l = \frac{1}{N}\sum_{j=1}^{N}
\max(0, 100 - 5 \cdot \sum_{f \in F_{lj}} v_f),
$$
where $F_{lj}$ is the set of the identified flaws for dimension $l$ on $r_{j}^{a}$, and $v_f$ is the severity level of flaw $f$. 
Additional details regarding the evaluation implementation are provided in Appendix \ref{appendix:LLMevaluation}.

\subsection{Alignment with Human judgment}
To assess whether our reference-augmented evaluation protocol is well-aligned with human judgment, we examine the correlation between human and LLM evaluator scores.
For evaluation scores averaged across the three dimensions, the alignment between human and LLM evaluator yields a Pearson's $r$ of $0.561$ and a Spearman's $\rho$ of $0.530$, indicating a moderate correlation by conventional social-science standards \citep{cohen2013statistical, lovakov2021empirically} and also consistent with the previous findings in role-playing evaluation \citep{zhou2025characterbench, he2025crab}.
These results demonstrate that our reference-augmented protocol serves as a reasonable proxy for human judgment.
Additional details of the correlation analysis are provided in Appendix \ref{appendix:interpretability_evaluation}, and the advantages of our protocol over traditional NLP metrics are reported in Appendix \ref{appendix:comparison_evaluation_metric}.

\section{Experimental Evaluation}
In this section, we present the experimental evaluation of SIBPersona. We first describe the process of building the dataset in \S \ref{sec:dataset_construction}, followed by the experimental settings in \S \ref{sec:experimental_setting}.
We then show the results of the experiments conducted on six influencers to demonstrate the effectiveness of SIBPersona in \S \ref{sec:results_of_llm_evaluations}.
To validate our SIBPersona, we perform a human evaluation to examine whether the $RP$ dimension provides interpretable evidence in \S \ref{sec:human_evaluations}. 
Lastly, we show that SIBPersona is also applicable to fictional character benchmarks in \S \ref{sec:transfer_to_fictional_character_benchmarks}.

\subsection{Dataset Construction}
\label{sec:dataset_construction}

We construct our dataset from posts, comments, and replies associated with six influencers, denoted as \textit{Influencer A} to \textit{F} from \textbf{CMoney Forum}\footnote{CMoney Forum is the largest online investor community in Taiwan, where members post Traditional Chinese messages under individual stock tickers and follow financial influencers whose posts routinely attract numerous replies.}.
We collect the data span from January 1, 2024, to September 30, 2025.
The data whose dates are before August 1, 2025 belong to the demonstration source, while the remaining data are preserved as the candidate pool for test samples.

To have a fair comparison between the different influencers, we perform stratified sampling on the candidate pool and keep the size of the test samples for each influencer in the same order of magnitude.
In addition, our preliminary analysis reveals that different comment–reply pairs exhibit distinct interaction types.
To ensure that the test samples thoroughly cover these diverse interactions, we prompt \texttt{gpt-4o-mini} to assign each comment–reply pair a score reflecting the degree of expressing personal characteristic.
Based on these scores, the data are partitioned into three levels: high, medium, and low.
From each level, we select $50$ test samples at most.
Additional details regarding dataset construction are provided in Appendix \ref{appendix:dataset}.

\subsection{Experimental Setting}
\label{sec:experimental_setting}

We compare our method with several ICL-based role-playing methods, all of which take the impersonated individual's profile $p$ as default input. 
We briefly describe each compared method below.
\textbf{Naive} relies solely on the impersonated individual's profile. 
\textbf{CoT} builds upon the \textbf{Naive} by additionally incorporating Chain-of-Thought reasoning \citep{wei2022chain}. 
\textbf{Vanilla RAG} augments the prompt with nine retrieved raw comment–reply pairs to match the number of dimensional documents utilized in SIBPersona.
Finally, \textbf{IMPersona} \citep{shi2025impersona} is a state-of-the-art ICL-based method for role-playing real individuals in conversations, which uses the proposed hierarchical memory module to model its data-variant persona.

We conduct two types of experiments to evaluate the robustness of SIBPersona along two axes: model choice and impersonated influencer. 
First, we examine model-agnostic robustness by applying multiple LLMs to construct data-variant personas, including \texttt{GPT-5.2}, \texttt{Gemini-3-flash}, \texttt{DeepSeek-v3.2}, and \texttt{Claude-Opus-4.6}. 
These experiments are conducted on \textit{Influencer A} to assess whether the effectiveness of SIBPersona is robust to the choice of the LLM used for persona construction, rather than being tied to a specific model. 
Second, we examine the robustness of cross-influencers by applying SIBPersona to \textit{Influencer B} through \textit{Influencer F}. 
In these experiments, we use \texttt{Gemini-3-flash} and \texttt{DeepSeek-v3.2} to evaluate whether SIBPersona remains effective across different influencers.
All experiments are evaluated using the reference-augmented evaluation protocol introduced in \S \ref{sec:evaluation_protocol} in all three dimensions, with \texttt{gpt-4.1} as the LLM evaluator and \texttt{text-embedding-3-small} as the embedding model for retrieval. For statistical significance testing, we perform t-tests with Holm correction for multiple comparisons.

\begin{table}[t!]
    \centering
    \setlength{\tabcolsep}{5pt}
    \begin{tabular}{l|cccc}
    \toprule
   \bf Methods & \bf $M$ & \bf $T$ & \bf $RP$ & \bf Avg.\\
      \midrule
IMPersona & 2.92$^\ddagger$ & 2.40$^\ddagger$ & 2.67$^\ddagger$  & 2.67$^\ddagger$ \\
Vanilla RAG & 3.68 & \bf 3.72 & \bf 3.62 & 3.67 \\
SIBPersona  & \bf 3.78 &  3.64 & \bf 3.62 & \bf 3.68 \\
\bottomrule
    \end{tabular}
    \caption{Human evaluation results comparing our framework with the baseline. \textbf{Bold}: better results; $^\ddag$: significant degradation compared to SIBPersona ($p<0.05$).}
    \label{table:human_evaluation_main}
\end{table}
\begin{table}[t!]
    \centering
    \setlength{\tabcolsep}{5pt}
    \begin{tabular}{l|ccc}
    \toprule
   \bf Conditions & \bf $M$ & \bf $T$ & \bf $RP$\\
    \midrule
Uninterpretable & 3.03 & 3.00 & 2.87  \\
Interpretable & \bf 4.13$^\dagger$ & \bf 3.97$^\dagger$ & \bf 4.00$^\dagger$ \\
\bottomrule
    \end{tabular}
    \caption{Human evaluation scores for Interpretable (n=20) and Uninterpretable (n=10) groups.  \textbf{Bold}: better results; $^\dag$: significant improvement with $p<0.05$.}
    \label{table:human_evaluation_interpret}
\end{table}

\subsection{Main Results}
\label{sec:results_of_llm_evaluations}
\autoref{table:main_model} shows the results of \textit{Influencer A} in four LLMs.
On average, SIBPersona outperforms all baselines by a statistically significant margin on both the $T$ and $RP$ dimensions.
Across all methods, performance on the $M$ dimension remains uniformly high, reflecting that replies on social media naturally demand less factual memory information and thus leave minimal room for penalties.
In terms of the other two dimensions, Naive and CoT perform the worst due to insufficient information regarding the impersonated individual.
While IMPersona incorporates rich memory information, it fails on speaking style and reaction process due to its single-dimensional design.
Vanilla RAG emerges as the strongest baseline, indicating that simply mirroring similar historical replies yields reasonable results.
Nevertheless, Vanilla RAG still falls short of SIBPersona, showing that constructing structured, multi-dimensional data-variant persona is crucial for faithful role-playing.

\autoref{table:main_people} reports the results aggregated over \textit{Influencer B} through \textit{Influencer F}. 
SIBPersona consistently achieves the highest scores in the $T$ and $RP$ dimensions across all five influencers under both models.
Nearly all baselines exhibit statistically significant degradation compared to SIBPersona, and the advantage is consistent regardless of model family or impersonated influencer.
These results show that SIBPersona is robust along two orthogonal axes: model choice and impersonated individual. 
From a qualitative perspective, \autoref{figure:good_case} illustrates that SIBPersona better captures the behavioral strategy and speaking style of the impersonated individual facing both praise and criticism.
Detailed experimental results and qualitative analysis are provided in Appendix \ref{appendix:perInfluencerResult} and \ref{appendix:CaseStudy}.

The ablation study further demonstrates the importance of the $RP$ dimension. 
As shown in \autoref{table:influencer_ablation}, removing the $RP$ dimension significantly degrades the performance on the $T$ and $RP$ dimensions, confirming that the $RP$ dimension contributes substantially to overall performance.
In contrast, removing only the internal state component causes marginal, non-significant declines. 
This asymmetry suggests that while the full SIB triplet is beneficial, explicit behavioral strategies $B$ contribute more critically than intermediate internal-state $I$ for the social media reply generation task, where replies tend to be short and direct.

\subsection{Human Evaluations}
\label{sec:human_evaluations}
To validate the effectiveness of SIBPersona, we perform a human evaluation in which the annotators rate the replies generated by SIBPersona, Vanilla RAG and IMPersona along three dimensions using a five-point Likert scale \citep{joshi2015likert}.
As shown in \autoref{table:human_evaluation_main}, SIBPersona significantly outperforms IMPersona across all three dimensions, while performing comparably to Vanilla RAG. Through qualitative analysis, we observe that Vanilla RAG frequently directly replicates the replies from the retrieved replies. This tendency often leads to misaligned replies when discrepancies arise between the follower's comment and the retrieved context. In contrast, SIBPersona generates replies grounded in the structured points, allowing it to adaptively fit the current situation. While such fine-grained differences are less distinguishable under the small-scale human evaluation, the adaptability of SIBPersona is more clearly reflected in our large-scale LLM-based evaluations.
Additional details and results of the human evaluation are provided in Appendix \ref{appendix:human_evaluation}. 

\paragraph{Interpretability of SIBPersona}
We further examine the interpretability of SIBPersona.
In this type of experiments, the annotators are asked to assess whether the SIB triplets in $\overline{PS_{dv}}$, especially the internal state and behavior strategy, can provide reasonable explanations for the generated reply given a test sample. 
Based on these assessments, we divide the test samples into interpretable and uninterpretable groups.
Additional details of the interpretability experiments are provided in Appendix \ref{appendix:interpretability_evaluation}.
As shown in \autoref{table:human_evaluation_interpret}, the replies in the interpretable group significantly outperform those in the uninterpretable group across all three dimensions.
This result suggests that the interpretability of the SIB triplets in $\overline{PS_{dv}}$ influences the quality of the generated replies and subsequently contributes to improved individual fidelity. 
In addition, we provide a qualitative example in Appendix~\ref{appendix:trace_example}, showing that SIBPersona can facilitate error diagnosis.

\subsection{Results on CharacterEval}
\label{sec:transfer_to_fictional_character_benchmarks}
To examine whether SIBPersona is applicable beyond the social media setting, we evaluate our method on CharacterEval \citep{tu2024charactereval}, a fictional character benchmark formulated as a single-turn response generation task. 
To minimize confounding effects from a model's prior knowledge of well-known fictional characters, we follow the anonymized evaluation setting proposed by \citet{peng2026rethinking}. 
We construct dimensional corpora using \texttt{gpt-4o-mini} and generate replies using \texttt{OLMo-3-7B-Instruct}.

As summarized in \autoref{table:charactereval_main}, SIBPersona significantly outperforms IMPersona on three out of four evaluation dimensions. Remarkably, SIBPersona also achieves performance competitive with \texttt{CoSER-Llama-3.1-8B} \citep{wang2025coser}, a model specifically fine-tuned on fictional characters' conversational data, demonstrating that an ICL-based method augmented with behavioral modeling can approach the performance of task-specific fine-tuned models.

\begin{table}[t!]
    \centering
    \setlength{\tabcolsep}{3pt}
    \begin{tabular}{l|cccc}
    \toprule
   \bf Methods & \bf CA & \bf RA & \bf CK & \bf CP \\
      \midrule
 IMPersona & 2.94 & 2.43 & 2.25 & 2.38 \\
SIBPersona & 3.28$^\dag$ & \bf 2.66$^\dag$ & \bf 2.32$^\dag$ & \bf 2.69$^\dag$ \\
\midrule
 \it Fine-tuned CoSER & \bf 3.35 & \it 2.61 & \it 2.28 & \it 2.68 \\
\bottomrule
    \end{tabular}
    \caption{Results on CharacterEval. \textbf{CA}: Conversational Ability, \textbf{RA}: Role-Playing Attractiveness, \textbf{CK}: Character Consistency–Knowledge, and \textbf{CP}: Character Consistency–Persona. \textbf{Bold}: the best results; $\dag$: significant improvement with $p<0.05$.}
    \label{table:charactereval_main}
\end{table}

The ablation study in \autoref{table:charactereval_ablation} further corroborates the findings in \S \ref{sec:results_of_llm_evaluations}.
Removing the $RP$ dimension leads to significant degradation across multiple evaluation dimensions, while removing only the internal state yields a smaller performance drop. 
Removing the $T$ dimension does not lead to a significant performance drop, suggesting that reaction process may play a more important role than stylistic information in fictional character benchmarks.
These consistent patterns across two distinct role-playing settings confirm that SIBPersona is broadly applicable beyond the domain for which it was originally designed.
Additional results and analysis are provided in Appendix \ref{appendix:RoleAgentBench}.

\begin{table}[t!]
    \centering
    \setlength{\tabcolsep}{5pt}
    \begin{tabular}{l|cccc}
    \toprule
   \bf Methods & \bf CA & \bf RA & \bf CK & \bf CP \\
      \midrule
SIBPersona &  3.28 &  2.66 & \bf 2.32 &  2.69 \\
 w/o $T$ & \bf 3.33 & \bf 2.69 & 2.31 & \bf 2.74 \\
  w/o $I$ & 3.28 & 2.66 & 2.31 & 2.68 \\
 w/o $RP$ & 3.16$^\ddagger$ & 2.61 & 2.30 & 2.62$^\ddagger$ \\
w/o $RP$ \& $T$ & 3.05$^\ddagger$ & 2.54$^\ddagger$ & 2.28 & 2.57$^\ddagger$ \\
\bottomrule
    \end{tabular}
    \caption{Component ablation study on CharacterEval. \textbf{Bold}: the best results; $\ddagger$: significant degradation compared to SIBPersona ($p<0.05$).}
    \label{table:charactereval_ablation}
\end{table}

\section{Conclusion}
In this work, we study the problem of role-playing real individuals on social media and propose a behavior-aware framework to address two major challenges.
In our framework, we introduce SIBPersona, which extends data-variant persona by modeling the reaction process.
We construct a dataset featuring real social media influencers and formulate the task of generating replies.
Our experimental results across multiple models demonstrate that SIBPersona improves role-playing fidelity, and human evaluation further verifies the interpretability of the retrieved SIB triplets.
We also find that reaction process modeling is applicable to fictional character role-playing benchmarks.
Moreover, our reference-augmented evaluation protocol for role-playing real-individual shows acceptable alignment with human judgment.
These findings suggest that behavior-aware persona modeling can improve role-playing performance on social media setting.

In future work, this framework could be extended to broader domains and more complex generation tasks. 
One such direction is extending SIBPersona to multi-turn dialogue, which would require handling evolving reaction processes.
This can be achieved by dynamically formulating retrieval queries from the accumulated conversational history instead of relying solely on the latest comment, enabling the model to fetch situation-relevant SIB triplets that capture accumulated context.

\section*{Limitations}
Although our study demonstrates effectiveness on both our newly constructed social media dataset and a public open-source dataset, and shows consistent improvements across both large API-based models and smaller open-source models, several limitations remain.

First, we observe that our method struggles with two types of comment situations (\autoref{figure:bad_case}): real-time information queries and sarcastic comments. In the former case, the LLM lacks up-to-date information, which leads to hallucinated replies. In the latter case, the LLM may misinterpret sarcastic remarks from followers, resulting in inappropriate behavioral strategies.

Second, our current study focuses on single-turn reply generation and does not evaluate performance in more challenging multi-turn dialogue settings. Since conversational contexts may evolve across turns, future work could investigate how the proposed framework performs under dynamically changing interaction scenarios.

Third, the dimensional documents in SIBPersona are currently provided to the LLM through in-context learning rather than model training. We have not explored whether integrating dimensional documents into model training could further improve performance. Future works could investigate training models with dimensional documents and examine whether such models can achieve stronger performance than API-based models.

\section*{Ethical Considerations}
This work involves the use of social media data and the impersonation of real individuals. 
We take several measures to address ethical concerns related to data usage, annotation practices, and potential societal impacts.

First, this study was conducted as an industry-academia collaboration between National Taiwan University and CMoney, the company that owns and operates the platform from which the data were collected. 
All data access and processing were conducted solely by CMoney-affiliated authors within a secure, company-managed environment in full compliance with the platform's Terms of Service and Privacy Policy.
We obtained explicit consent from all influencers whose personas were evaluated in our experiments to use their available data for research purposes.
Second, regarding follower comments, platform members accept the Terms of Service and Privacy Policy upon registration, authorizing the platform to collect, process, and use member-provided data for online behavior research, statistics, and service improvements.
To protect participant privacy, all follower comments in our dataset and evaluation materials are strictly anonymized, retaining no user identifiers or personally identifiable information.
Furthermore, no raw influencer or follower content will be released as part of our research artifacts.
Third, human annotators involved in the evaluation are employees of CMoney and are compensated through their regular salaries.
Fourth, AI-based writing assistance tools were used to help polish parts of the manuscript.
All AI-assisted text was carefully reviewed and proofread by the authors to ensure correctness and accuracy.

Finally, we acknowledge that technologies capable of impersonating real individuals could potentially be misused to generate misleading or deceptive content on social media.
This work is an evaluative research study rather than a public deployment.
Any future commercialization or deployment would be handled separately by CMoney and is outside the scope of this study;
nonetheless, we recommend that any downstream deployment require: (1) personas to be created internally and only for explicitly authorized influencers, (2) human review of generated content before publication, particularly for sensitive responses, and (3) mechanisms allowing influencers to deactivate their personas or withdraw consent. 
Meanwhile, our study contributes to understanding the capabilities and limitations of such systems. We hope that these insights can help future research develop methods for detecting impersonation and mitigating the spread of misinformation in online communities.

\section*{Acknowledgments}
The work was financially supported by a National Taiwan University Industry-Academia Cooperative Research Project, the National Science and Technology Council (NSTC), and the Featured Area Research Center Program within the framework of the Higher Education Sprout Project by the Ministry of Education (MOE),  Taiwan, under Grants 112-2223-E-002-012-MY5, 115-2628-E-002-023-MY4, and 115L900901.

We thank Li-Ching Chien, Pei-Yu Hou, Pei-Ru Huang, Si-Rui Huang, Meng-Jung Lin, Yung-Chen Liu, Xiao-An Wang, and Yi-Chi Yeh for serving as human annotators in this work.

\bibliography{custom}

\appendix
\section{Detail of Constructing Dataset}
\label{appendix:dataset}
In this work, we construct the dataset from social media data related to the impersonated influencers, collected internally under the authorization of the platform operator. To ensure sufficient data coverage, we select influencers who have more than $500$ comment--reply pairs in the demonstration source and more than $120$ comment--reply pairs in the candidate pool for testing. \autoref{table:raw_data} reports the dataset statistics for the six influencers studied in our work.

\begin{figure*}[hbt!]
\centering
    \includegraphics[width=\textwidth]{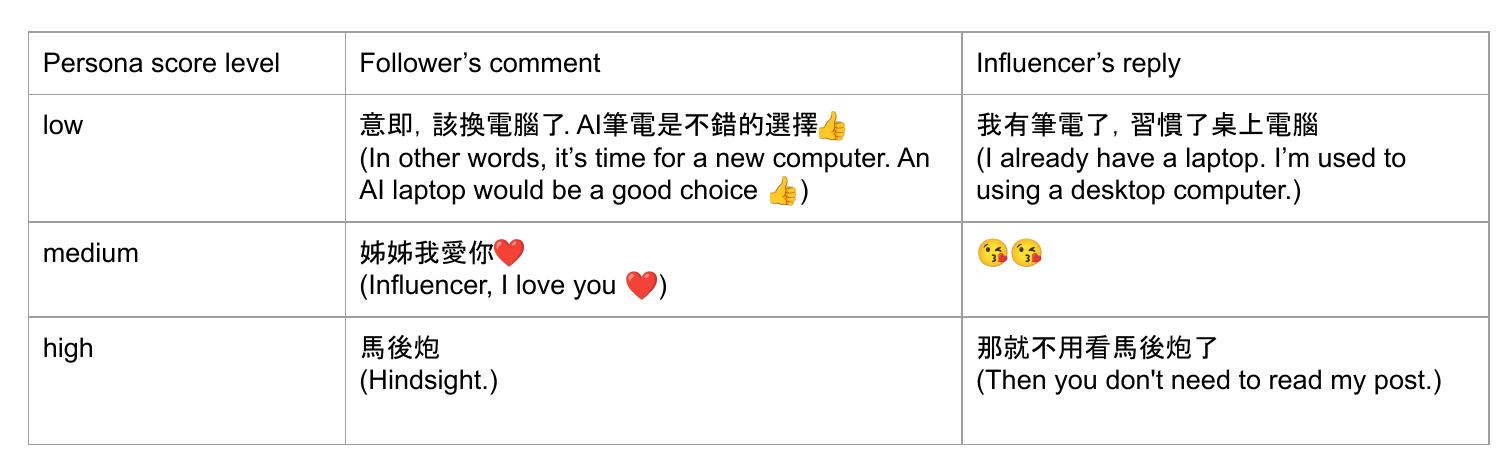}
    \caption{Examples of three level of data for \textit{Influencer A}.}
    \label{figure:persona_score_data}
\end{figure*}

Furthermore, based on our preliminary analysis, comment--reply pairs exhibit highly distinct interaction types. 
Some pairs mainly involve factual information exchange, while others more clearly reveal the influencer's speaking style or behavioral strategies. 
To ensure that the test samples comprehensively covers these diverse interactions, we prompt \texttt{gpt-4o-mini} to assign each comment--reply pair a persona score ranging from 0 to 1, indicating the degree to which the reply exhibits the influencer's personal characteristic. 
We then use this score as the basis for stratified sampling. 
Specifically, comment--reply pairs are categorized into three levels according to their persona scores: high $[0.8, 1.0]$, medium $[0.5, 0.8)$, and low $[0, 0.5)$. 
Examples of the three levels are shown in \autoref{figure:persona_score_data}. 
Low-level pairs mainly involve factual or experiential information exchange, medium-level pairs exhibit stylistic cues such as emoji usage, and high-level pairs clearly reflect the influencer's personal characteristic. 
The prompt used to obtain persona scores is shown in \autoref{figure:persona_score_prompt}. 
The statistics of the test samples for each of the six influencers are summarized in \autoref{table:dataset}. 
For \textit{Influencer D}, all available data in the medium and high levels are included because each level contains fewer than $50$ test samples. 
The same logic applies to \textit{Influencer E} and \textit{Influencer F}. 

To further demonstrate the diversity of our dataset, \autoref{table:dataset_diversity} reports the text length distributions and the proportions of the three persona-score levels across all six influencers. 
As shown in the table, textual lengths vary considerably across individuals: average reply lengths range from $8.46$ to $33.65$ characters, while average post lengths range from $271.65$ to $952.74$ characters. 
Moreover, the distribution of persona-score levels in the candidate pool reflects distinct individual interaction styles.
For instance, \textit{Influencer E}'s comment--reply pairs predominantly involve low persona-characteristic exchanges ($87\%$), whereas \textit{Influencer C} displays a much more balanced distribution across medium ($36\%$) and low ($53\%$) levels. These statistics confirm that our dataset encompasses a wide spectrum of behavioral patterns.

\begin{table}[t!]
    \centering
    \setlength{\tabcolsep}{5pt}
    \begin{tabular}{l|cccccc}
    \toprule
        & \multicolumn{6}{c}{\textit{Influencer}} \\
   \bf Persona Score  & \it A & \it B & \it C & \it D & \it E & \it F\\
      \midrule
High  & 50 & 50 & 50 & 35 & 50 & 9\\
Medium  & 50 & 50 & 50 & 50 & 22 & 39\\
Low & 50 & 50 & 50 & 50 & 12 &  50\\
\bottomrule
    \end{tabular}
    \caption{Number of data in the test samples.}
    \label{table:dataset}
\end{table}
\begin{table*}[hbt!]
\centering
\setlength{\tabcolsep}{3pt}
\begin{tabular}{lcccccc}
\toprule
\textbf{Influencer} & \textbf{Post Length} & \textbf{Comment Length} & \textbf{Reply Length}  & \textbf{High (\%)} & \textbf{Med (\%)} & \textbf{Low (\%)} \\
\midrule
\textit{A} & 467.92 $\pm$ 471.09  & 17.58 $\pm$ 25.37 & 8.46 $\pm$ 9.69  & 13\% & 29\% & 58\% \\
\textit{B} & 367.39 $\pm$ 231.62  & 30.70 $\pm$ 57.55& 10.01 $\pm$ 10.54  & 12\% & 35\% & 53\% \\
\textit{C} & 952.74 $\pm$ 740.92  & 66.83 $\pm$ 105.54& 33.65 $\pm$ 28.14 & 34\% & 36\% & 53\% \\
\textit{D} & 355.21 $\pm$ 508.89   & 14.20 $\pm$ 15.24& 9.56 $\pm$ 10.17  & 16\% & 25\% & 59\% \\
\textit{E} & 271.65 $\pm$ 221.00   & 27.54 $\pm$ 28.06& 9.52 $\pm$ 10.21  & 5\% & 8\%  & 87\% \\
\textit{F} & 860.21 $\pm$ 574.41  & 26.53 $\pm$ 29.41& 11.82 $\pm$ 10.39  & 8\% & 33\% & 60\% \\
\bottomrule
\end{tabular}
\caption{Dataset diversity statistics across the six influencers, including text lengths and the proportion of comment--reply pairs across high, medium, and low persona-score levels.}
\label{table:dataset_diversity}
\end{table*}

\section{Detail of SIBPersona}
\label{appendix:SIBPersona}
\subsection{Data-invariant Persona Construction}
For the data-invariant persona, we construct a short profile for each influencer using the influencer description provided on the social media, rather than deriving it from historical data. Each profile contains approximately $50$--$100$ words.

\subsection{Data-variant Persona Construction}
\label{appendix:persona_bank}
\autoref{table:persona} reports the size of the dimensional corpora extracted for \textit{Influencer A} in four LLMs, while \autoref{table:persona_people_deepseek} and \autoref{table:persona_people_gemini} further present the corresponding statistics for the remaining five influencers using \texttt{DeepSeek-v3.2} and \texttt{gemini-3-flash-preview}, respectively.
All models are accessed through cloud-based APIs: we access \texttt{claude-opus-4-6} (pinned snapshot) through the Claude API, \texttt{gemini-3-flash-preview} (December 2025 snapshot) through Google Cloud Platform (GCP), and both \texttt{gpt-5.2-2025-12-11} and \texttt{DeepSeek-v3.2} through Azure AI Foundry.
For decoding parameters across tasks, we set $(\text{temperature}, \text{top-}p)$ to $(1.0, 1.0)$ for recognizing, $(0.2, 0.8)$ for extraction, and $(0.8, 1.0)$ for generation. All other API parameters remain at their default values, with the reasoning effort set to the lowest level supported by each model.

\subsection{Distinctiveness of Persona Dimensions}
\label{appendix:persona_distinct}
Our dimension-wise retrieval design is based on the hypothesis that the three dimensions of data-variant persona capture largely distinct aspects of an individual's persona.
While these dimensions may still exhibit certain correlations, we hypothesize that memory, speaking style, and reaction process encode sufficiently different types of information to be treated separately during retrieval.
Under this hypothesis, the follower's comment is matched separately against each dimensional corpus, which prevents high-frequency patterns in one dimension from dominating the retrieval results of the others.

To empirically examine this hypothesis, we analyze the distinctiveness of the three dimensions in the embedding space.
Specifically, we embed $m_{p_jk}$ for memory, $sp_{p_jk}$ for speaking style, and behavior $b_{jk}$ for the reaction process.
We use $b_{jk}$ rather than the full SIB triplet because behavior, similar to memory and speaking style, directly reflects the impersonated individual's observable reply patterns, whereas situation and internal state mainly describe the intermediate reasoning process that leads to the behavior.
The resulting vectors are then analyzed through qualitative visualization and quantitative clustering evaluation.
The results provide empirical evidence that the three dimensions are separable, supporting our decision to process them independently during retrieval.

\begin{figure}[hbt!]
\centering
    \includegraphics[width=0.45\textwidth]{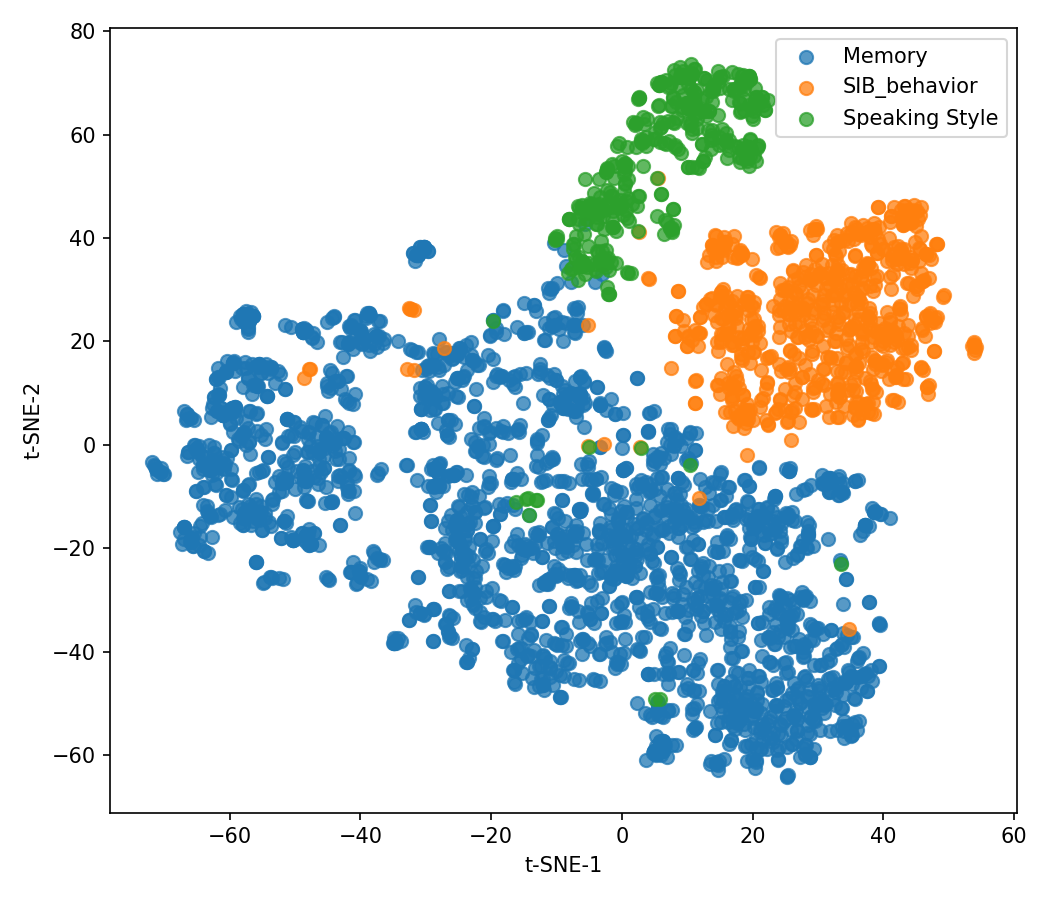}
    \caption{t-SNE projection of the embeddings of $mp_{jk}$, $sp_{jk}$, and $b_{jk}$ generated by \texttt{Claude-Opus-4.6}.}
    \label{figure:tsne_claude_plot}
\end{figure}

For the qualitative analysis, \autoref{figure:tsne_claude_plot} shows the t-SNE visualization of the embeddings of $mp_{jk}$, $sp_{jk}$, and $b_{jk}$ extracted for \textit{Influencer A} by \texttt{Claude Opus 4.6}. As illustrated in the figure, the three dimensions occupy different regions in the projected space, suggesting that they capture different aspects of the impersonated individual. For the quantitative analysis, we further calculate the Silhouette Score across six influencers, using documents extracted by \texttt{DeepSeek-v3.2}. The resulting average Silhouette Score is $0.37$. As a random baseline, we permute the cluster label of all document embeddings and repeat this process 1,000 times. The average Silhouette Score under this random permutation setting is $-0.001$. The substantial gap between the actual score and the random baseline provides evidence that the three dimensions form distinguishable clusters in the embedding space.

\begin{table}[t!]
    \centering
    \setlength{\tabcolsep}{5pt}
    \begin{tabular}{l|ccc}
    \toprule
   Models  & \bf $M$ & \bf $T$ & \bf $RP$\\
      \midrule
\texttt{GPT-5.2}  & 6,064 & 1,410 & 970 \\
\texttt{Gemini-3-flash}  & 5,755 & 1,050 & 861 \\
\texttt{Deepseek-V3.2}  & 5,143 & 1,022 & 749 \\
\texttt{Claude-Opus-4.6}  & 2,307 & ~~407 & 703 \\
\bottomrule
    \end{tabular}
    \caption{Number of dimensional documents of \textit{Influencer A}.}
    \label{table:persona}
\end{table}
\begin{table}[t!]
    \centering
    \setlength{\tabcolsep}{5pt}
    \begin{tabular}{c|ccc}
    \toprule
   \textbf{Influencer}  & \bf $M$ & \bf $T$ & \bf $RP$\\
      \midrule
\textit{B}  & 12,358  & 8,162  & 6,701  \\
    \textit{C}  & ~~6,179  & 2,982  & 4,049  \\
    \textit{D}  & ~~8,914  & 4,935  & 3,811  \\
    \textit{E}  & 10,087  & 2,393  & 2,378  \\
    \textit{F}  & ~~2,587  & ~~446  & ~~398  \\
\bottomrule
    \end{tabular}
    \caption{Number of dimensional documents of \textit{Influencer B-F} extracted from \texttt{Deepseek V3.2}.}
    \label{table:persona_people_deepseek}
\end{table}
\begin{table}[t!]
    \centering
    \setlength{\tabcolsep}{5pt}
    \begin{tabular}{c|ccc}
    \toprule
   \textbf{Influencer}  & \bf $M$ & \bf $T$ & \bf $RP$\\
      \midrule
\textit{B}  & 13360  & 8253  & 6911  \\
\textit{C}  & 5296  & 3321  & 3643  \\
\textit{D}  & 10102  & 4839  &  4053 \\
\textit{E}  & 10309  & 3536  & 3287  \\
\textit{F}  & 2363  & 515  &  449 \\
\bottomrule
    \end{tabular}
    \caption{Number of dimensional documents of \textit{Influencer B-F} extracted from \texttt{Gemini-3-flash}.}
    \label{table:persona_people_gemini}
\end{table}

\section{Details of Reference-Augmented Evaluation}
\label{appendix:LLMevaluation}
We adopt the penalty-based evaluation protocol introduced by \citet{wang2025coser} to score RPA-generated replies.
We use \texttt{gpt-4.1-2025-04-14} as LLM evaluator and access it through the Azure AI Foundry. The temperature is set to $0.2$ and top-$p$ to $0.8$ to ensure the stability of the evaluation results.

\paragraph{Number of Representative Dimensional Documents}
To avoid using the information from the same dimensional corpora for RPA generation and evaluation, we select the representative dimensional documents from the dimensional corpora constructed from the candidate pool for test samples. Following the rule of thumb discussed by \citet{royall2015finding}, we set the number of medoids $k$ according to $k=\sqrt{M/2}$, where $M$ denotes the number of dimensional documents in the corresponding corpus.
We then round $k$ up to the nearest integer and use it as the number of representative dimensional documents. \autoref{table:document_statistics} reports, for each influencer and extracting model, the total number of extracted dimensional documents and the corresponding number of representative dimensional documents used in evaluation.

\paragraph{Rubrics for LLM Evaluators}
Evaluations are conducted along three dimensions: memory, speaking style, and reaction process.
Each dimension is independently evaluated in a separate run.
For each evaluation run, the LLM evaluator is provided with the definition of the target dimension and its corresponding error types as dimension-specific rubrics.
Using the data-invariant profile and representative dimensional documents as references, the LLM evaluator identifies flaws in the generated reply and assigns a severity score to each detected flaw.
This design allows the LLM evaluator to focus on one dimension at a time and avoids penalizing errors from other dimensions.
The complete evaluation prompt and evaluation rubrics are shown in \autoref{figure:evaluation_prompt} and \autoref{figure:evaluation_rubrics}, respectively.

\paragraph{Details of the Formula}
In the original formulation by \citet{wang2025coser}, the evaluation score includes a length-based compensation term in addition to the penalty term, since longer replies are more likely to receive penalties under error-based scoring schemes.
However, in our setting, the generated replies are typically short, with an average length of approximately $10$ words.
We therefore omit the length-based score adjustment in our evaluation.

\begin{figure}[t!]
\centering
    \includegraphics[width=0.45\textwidth]{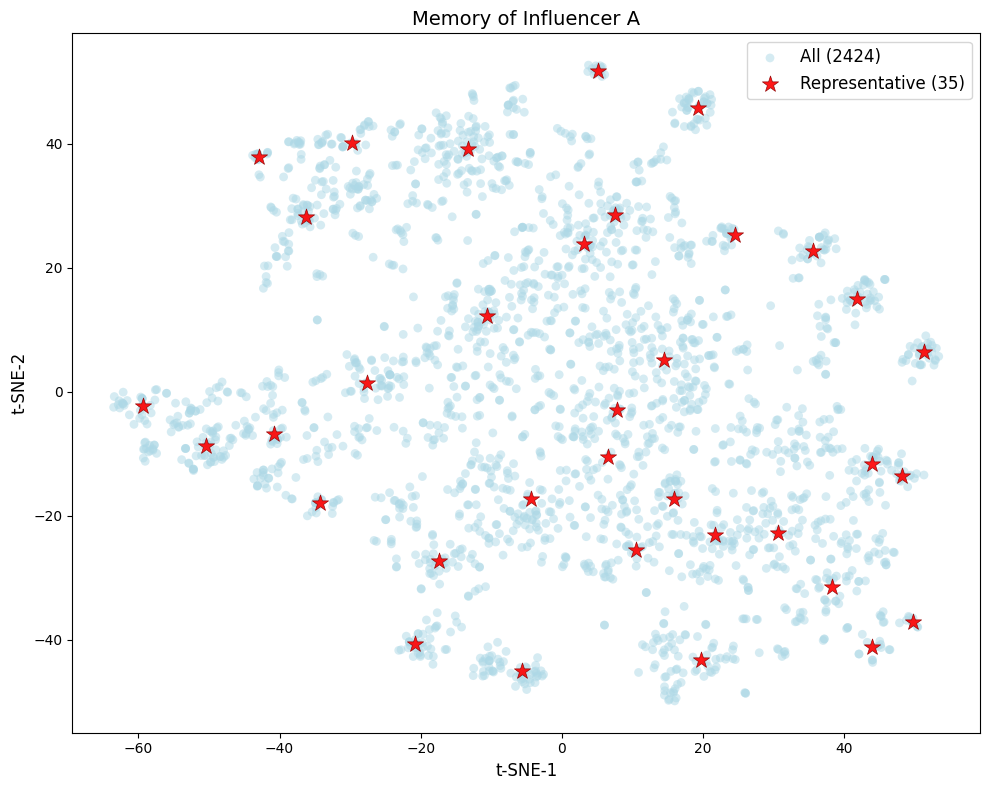}
    \caption{t-SNE projection of representative memory documents generated by \texttt{DeepSeek-v3.2}.}
    \label{figure:rep_memory}
\end{figure}

\begin{figure}[hbt!]
\centering
    \includegraphics[width=0.45\textwidth]{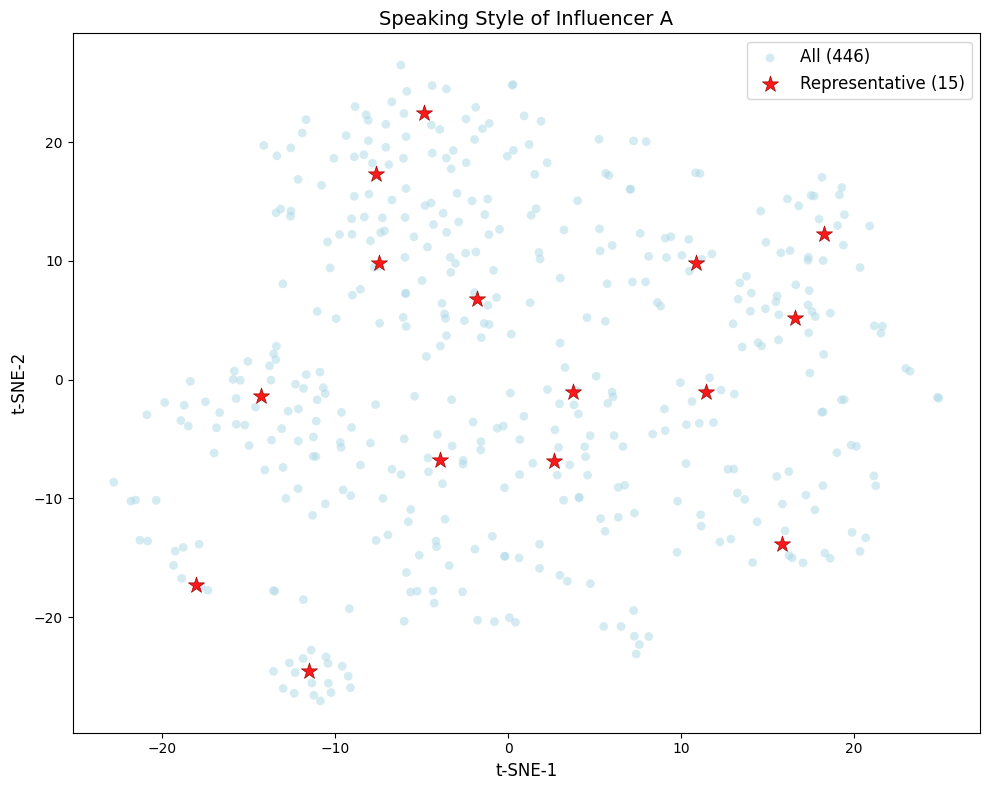}
    \caption{t-SNE projection of representative speaking style documents generated by \texttt{DeepSeek-v3.2}.}
    \label{figure:rep_speaking}
\end{figure}

\begin{figure}[hbt!]
\centering
    \includegraphics[width=0.45\textwidth]{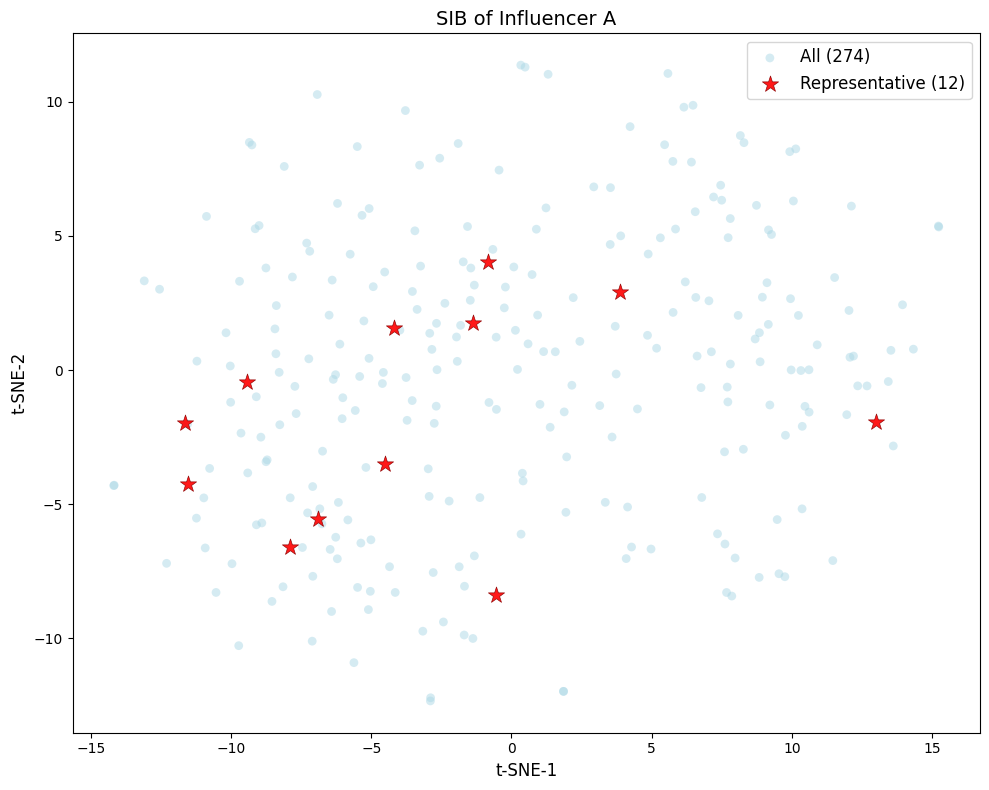}
    \caption{t-SNE projection of representative reaction process documents generated by \texttt{DeepSeek-v3.2}.}
    \label{figure:rep_SIB}
\end{figure}

\paragraph{Visualization for Representative Dimensional Documents}
We visualize the representative dimensional documents extracted for \textit{Influencer A} by \texttt{DeepSeek-v3.2}. As shown in \autoref{figure:rep_memory}, \autoref{figure:rep_speaking}, and \autoref{figure:rep_SIB}, the selected representative dimensional documents are spread across different regions of the projected space rather than being concentrated in a single local cluster.
This qualitative observation suggests that representative dimensional documents capture diverse aspects of the impersonated influencer across the three dimensions.

\section{Details of Human Evaluation}
\label{appendix:human_evaluation}
Human evaluations are conducted with two main objectives. 
The first objective is to assess the alignment between reference-augmented evaluation protocol and human judgment.
For this purpose, we use the comparison evaluation to support annotator quality control and the performance evaluation to measure the correlation between human judgment and LLM evaluators. 
The second objective is to examine the validity of the reaction-process-based design of SIBPersona. 
Specifically, the interpretability evaluation assesses whether the retrieved SIB triplets can serve as explanations for generated replies, and the construction faithfulness evaluation verifies whether the constructed SIB triplets are faithful to their corresponding evidence pairs. 
The following subsections first describe the settings and annotation protocols of the four human evaluations in Appendix \ref{appendix:human_evaluation_setting} and then present the corresponding results in Appendix \ref{appendix:human_evaluation_results}.

\subsection{Human Evaluation Setting}
\label{appendix:human_evaluation_setting}
This subsection describes the setting of our human evaluation. In the following, we refer to each question in the human evaluation inventory as an evaluation item.
\subsubsection{Comparison Evaluation}
The comparison evaluation is designed to serve two complementary purposes. 
First, it provides a post hoc diagnostic for identifying potentially unreliable annotators. 
Second, it offers a comparison-based assessment of the quality of RPA-generated replies. 
For each evaluation item, annotators are given an influencer's post $ip$, a follower's comment $fc$, and three anonymized candidate replies. 
They are asked to rank the three replies according to how closely each reply resembles the impersonated influencer's actual reply, with ties allowed when multiple replies are considered equally similar.

Although all three options are presented as generated replies, only two of them are produced by RPA: one by IMPersona and the other by SIBPersona. 
The remaining option is the ground-truth reply written by the influencer. 
Since the ground-truth reply is written by impersonated individual, a reliable annotator is expected not to consistently rank it as the least similar reply. 
We therefore use the hidden ground-truth reply as a quality-control signal. 
Specifically, for each annotator, we compute the proportion of evaluation items in which the ground-truth reply is ranked last. 
Annotators whose ground-truth-last rate exceeds a predefined threshold are excluded from subsequent analysis; in our work, the threshold is set to $30\%$. 

\subsubsection{Performance Evaluation}
\label{appendix:performance_evaluation_setting}
The performance evaluation is designed to assess the alignment between our proposed evaluation protocol and human judgment by measuring the correlation between their scores.
In addition, it allows us to compare the quality of RPA-generated replies through statistical significance testing.
For each evaluation item, annotators are given an influencer's post, a follower's comment, and an RPA-generated reply generated by IMPersona, Vanilla RAG or SIBPersona. 
They are asked to rate the reply along three dimensions: memory, speaking style, and reaction process, using a five-point Likert scale \citep{joshi2015likert}.
If an annotator considers a reply irrelevant to a given dimension, they may select \textit{NA}.
This option is included to avoid forcing annotators to assign scores to dimensions that are not applicable to a particular reply.

To address \textit{NA} annotations during analysis, we process human evaluation scores as follows.
For each evaluation item and each dimension, we aggregate scores across annotators by averaging all non-\textit{NA} ratings.
This aggregation yields one human evaluation score for each dimension of each sample and helps reduce the influence of annotator-specific variation, allowing the subsequent correlation analysis to focus on aggregate human evaluation trends. 
If all annotators select \textit{NA} for an evaluation item on a given dimension, the aggregated score for that dimension is treated as \textit{NA}.
Such cases are excluded from analysis involving the corresponding dimension because no valid human evaluation score is available.

After aggregating the human evaluation scores, we perform two complementary analysis. 
The first analysis examines the alignment between LLM evaluator and human judgment using correlation measures. 
The second analysis compares the quality of RPA-generated replies across IMPersona, Vanilla RAG and SIBPersona based on human evaluation scores.

For the correlation analysis, we compute both dimension-specific and overall correlations between LLM evaluator and human judgment scores.
Dimension-specific correlation is calculated separately for memory, speaking style, and reaction process. 
For each dimension, evaluation items with \textit{NA} in the corresponding aggregated human judgment score are excluded since no valid human score is available for that dimension.
This analysis allows us to examine whether the LLM evaluator aligns with human judgment on each specific dimension. 

In addition, we compute an overall correlation by first averaging the scores across the three dimensions for each evaluation item and then calculating the correlation based on these overall scores. 
Compared with the dimension-specific correlation, this analysis provides an overall view of the alignment between LLM evaluator and human judgment. 
For this analysis, evaluation items are excluded if any of the three dimension scores is \textit{NA}, because the overall score cannot be consistently computed across all dimensions.

For the method-level quality comparison, we focus on the three RPA methods: IMPersona, Vanilla RAG and SIBPersona. 
For each evaluation dimension, we compute the average human evaluation score of each method and conduct $t$-tests with Holm correction to examine whether the score differences are statistically significant. 
This analysis complements the correlation analysis by directly comparing the human-evaluated quality of the three methods.

\subsubsection{Interpretability Evaluation}
After evaluating the quality of the RPA-generated replies, we further conduct evaluations to examine whether the constructed reaction process corpus provides meaningful explanatory evidence for SIBPersona.
We first perform an interpretability evaluation to assess whether the retrieved SIB triplets can serve as plausible explanations for the generated replies. 
For each evaluation item, annotators are given an influencer's post, a follower's comment, and a reply generated by SIBPersona, together with the top-three SIB triplets retrieved and provided to the RPA. 

Annotators are asked to evaluate each SIB triplet from two aspects. 
First, they assess whether the situation is relevant to the influencer's post and the follower's comment. 
Second, they assess whether the internal state and behavior can reasonably explain the generated reply. 
This design allows us to examine both the contextual relevance of the retrieved situations and the explanatory consistency between the retrieved SIB triplets and the final generated reply. 
Ratings are given using a five-point Likert scale. 

\subsubsection{Construction Faithfulness Evaluation}
Finally, we conduct a construction faithfulness evaluation to assess whether the generated SIB triplets are reasonably extracted from the original data sample.
This evaluation examines whether each extracted SIB triplet can be reasonably supported by the comment--reply pair assigned as its evidence.
For each evaluation item, annotators are given a comment--reply pair and its corresponding extracted SIB triplet. 
They are then asked to the annotator whether it is reasonable to construct the given SIB triplet from the provided comment--reply pair. 
This setup allows us to directly evaluate the faithfulness of the corpus construction process, rather than only assessing their groundedness in downstream reply generation. 
The evaluation is collected as a binary pass/fail label, where \textit{pass} indicates that the triplet is reasonably supported by the source comment--reply pair, and \textit{fail} indicates that the triplet is unsupported, inconsistent with the comment--reply pair, or overly inferred from the source data. 

\subsubsection{Implementation Details}
All human evaluations are conducted using test samples generated by \texttt{DeepSeek-V3.2} for \textit{Influencer F}. 
For the performance evaluation, we use $30$ test samples and evaluate replies generated by three methods, Vanilla RAG, IMPersona and SIBPersona, resulting in $90$ evaluation items in total.
For the interpretability evaluation, we use the same $30$ test samples. 
Since each test sample contains top-three retrieved SIB triplets, annotators evaluate each triplet separately.

The $30$ test samples are manually selected from the three levels introduced in \S \ref{sec:dataset_construction} to ensure diversity in comment situations.
For the comparison evaluation, we randomly select $10$ samples from these $30$ test samples.
For the construction faithfulness evaluation, we use $50$ representative dimensional documents from reaction process corpus to assess whether the extracted SIB triplets are faithful to their corresponding evidence pairs.

We recruit nine annotators for the human evaluation. 
All annotators have worked with \textit{Influencer F} for several years, ensuring that they have sufficient familiarity with the impersonated influencer. 
To control annotation quality, we further filter annotators based on the ground-truth-last rate calculated from the comparison evaluation. 
Since only three annotators pass the filtering threshold, we report the results of these three annotators in the following subsection.

\subsection{Results of Human Evaluation}
\label{appendix:human_evaluation_results}

\paragraph{Results of Comparison Evaluation}
\label{appendix:comparison_evaluation}
For the comparison evaluation, we convert the rankings of candidate replies into reciprocal ranking scores, where the score of each reply is computed as the reciprocal of its assigned rank. 
We then report the average reciprocal ranking score for each method, with higher scores indicating better rankings. 
As shown in \autoref{table:human_comparison}, SIBPersona achieves a significantly higher score than IMPersona and obtains an average score comparable to that of the ground-truth replies. 
This result suggests that annotators judged the replies generated by SIBPersona to be substantially closer to the replies of real influencer's than those generated by IMPersona.

\begin{table}[t!]
    \centering
    \setlength{\tabcolsep}{5pt}
    \begin{tabular}{l|c}
    \toprule
   \bf Methods & \bf Reciprocal Ranking score\\
      \midrule
IMPersona & 0.42$^\ddagger$ \\
SIBPersona  & \bf 0.71\\
Ground Truth & \bf 0.71\\
\bottomrule
    \end{tabular}
    \caption{Human comparison evaluation reciprocal scores. \textbf{Bold} indicates the best score. $\ddagger$: significant degradation compared to SIBPersona ($p<0.05$)}
    \label{table:human_comparison}
\end{table}

\begin{figure*}[hbt!]
\centering
    \includegraphics[width=\textwidth]{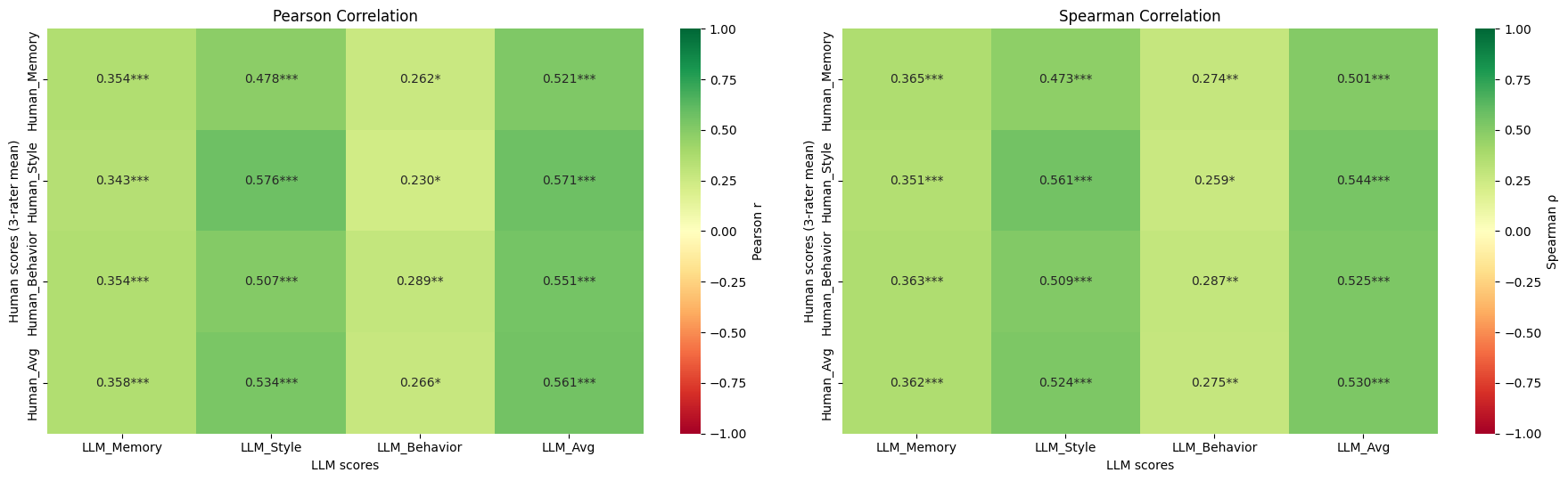}
    \caption{Correlation matrix between LLM and human evaluations. *** $p<0.001$, ** $p<0.01$, * $p<0.05$.}
    \label{figure:correlation_plot}
\end{figure*}

\paragraph{Results of Performance Evaluation}
To examine the alignment between human judgment and LLM evaluators, we compare the aggregated human judgment scores with the LLM evaluator scores and report their correlations. 
We report both Pearson's $r$ \citep{pearson1920notes} and Spearman's $\rho$ \citep{ca468a70-0be4-389a-b0b9-5dd1ff52b33f} to measure linear and rank-based associations, respectively.
As shown in \autoref{figure:correlation_plot}, the overall human judgment score shows a moderate correlation with the overall LLM evaluator score, yielding Pearson's $r$ of $0.561$ and Spearman's $\rho$ of $0.530$.
This suggests that the LLM evaluator captures evaluation trends that are broadly aligned with human judgment when the three dimensions are considered jointly. 
Among the dimension-specific correlations, speaking style shows the strongest alignment between human judgment and LLM evaluator, with Pearson's $r$ of $0.576$ and Spearman's $\rho$ of $0.561$ indicating a moderate correlation.
Memory and reaction process also show significant but weaker correlations, with Pearson's $r$ of $0.354$ and $0.289$, and Spearman's $\rho$ of $0.365$ and $0.287$, respectively.
These results are consistent with the previous findings in role-playing evaluation \citep{zhou2025characterbench, he2025crab} and demonstrate that our reference-augmented protocol serves as a reasonable proxy for human judgment.

Moreover, we further assess inter-rater agreement among human annotators, computed by the weighted Cohen's $\kappa$ \citep{warrens2015five}. 
The agreement scores across the three dimensions are presented in \autoref{table:inter_rater}. 
The pairwise agreement scores range from $0.364$ to $0.508$. 
Following the interpretation guidelines of \citet{mchugh2012interrater}, these results indicate a fair to moderate level of agreement across annotators. 
The agreement is highest between Rater 1 and Rater 3, while Rater 1 and Rater 2 show relatively lower agreement. 
This pattern suggests that the annotation task involves subjective judgment, but the annotators still exhibit meaningful consistency. 

For the method-level quality comparison, as shown in \autoref{table:human_evaluation_main}, SIBPersona significantly outperforms IMPersona across all three dimensions.

\begin{table}[t!]
    \centering
    \begin{tabular}{c|ccc}
    \toprule
        & \bf $M$ & \bf $T$ & \bf $RP$ \\
      \midrule
    \bf Rater 1 \& 2 & 0.364 & 0.429 & 0.406  \\
    \bf Rater 1 \& 3 & 0.417 & 0.506 & 0.508  \\
    \bf Rater 2 \& 3  & 0.426 & 0.495 & 0.487  \\
\bottomrule
    \end{tabular}
    \caption{Inter-rater agreement among the three annotators.}
    \label{table:inter_rater}
\end{table}

\paragraph{Results of Interpretability Evaluation}
\label{appendix:interpretability_evaluation}
For the interpretability evaluation, each test sample is associated with three retrieved SIB triplets. Since the RPA may rely more heavily on one of the retrieved triplets during generation, we take the highest score among the three triplets as the score for that test sample. Specifically, let $score^{S}_{i,j}$ and $score^{IB}_{i,j}$ denote the average annotator scores for the $j$-th SIB triplet of the $i$-th test sample, where $j \in \{1,2,3\}$. The final scores, denoted as $Score_S$ and $Score_{IB}$, are computed by averaging the highest triplet-level score across all $N$ test samples:
\begin{align*}
Score_S = \frac{1}{N} \sum_{i=1}^{N} \max_{j \in \{1,2,3\}} score^{S}_{i,j},
\end{align*}
\begin{align*}
Score_{IB} = \frac{1}{N} \sum_{i=1}^{N} \max_{j \in \{1,2,3\}} score^{IB}_{i,j}.
\end{align*}

Our RPA replies achieve a $Score_S$ of $3.63$ and a $Score_{IB}$ of $3.57$.
The high $Score_S$ indicates that the situation described in the retrieved SIB triplet generally align with the corresponding follower's comment.
The high $Score_{IB}$ further suggests that the internal states and behaviors described in the triplets can reasonably explain the final replies generated by the RPA.

In addition, we use $Score_{IB}$ as the criterion for categorizing test samples into interpretable and uninterpretable, as shown in \autoref{table:human_evaluation_interpret}.
For each test sample, if the average $Score_{IB}$ assigned by the three annotators exceeds $3$, the test sample is considered interpretable; otherwise, it is considered uninterpretable.
These results suggest that our retrieval method can identify situations similar to an unseen follower's comment, and the retrieved internal states and behaviors serve as a valid supporting context in the RPA-generated replies.

\paragraph{Results of Extraction Faithfulness Evaluation}
\label{appendix:extraction_quality_evaluation}
To evaluate the faithfulness of the reaction process corpus, we report the pass rates assigned by the three annotators in \autoref{table:extraction_quality}. 
The individual annotator pass rates are consistently high, with all raters assigning pass rates above $0.80$. 
When applying majority voting across annotators, the overall pass rate reaches $0.94$ with a 95\% confidence interval of $[0.84, 0.98]$.

These results suggest that most constructed SIB triplets are reasonably supported by their corresponding comment--reply pair. 
In other words, our process of constructing dimensional corpora can generally preserve information from the original comment--reply pairs while converting them into structured SIB triplets. 
The higher majority-vote pass rate further indicates that although the annotators may differ in their strictness, there is strong aggregate support for the faithfulness of the constructed reaction process corpus.

\begin{table}[t!]
    \centering
    \begin{tabular}{c|cc}
    \toprule
        & \bf Pass Rate & \bf 95\% CI  \\
      \midrule
    \bf Rater 1 & 0.80 &  [0.67, 0.89]  \\
    \bf Rater 2 & 0.84 &  [0.72, 0.92] \\
    \bf Rater 3  & 0.92 &  [0.81, 0.97]  \\
    \midrule
    \bf Total & 0.94 & [0.84, 0.98] \\
\bottomrule
    \end{tabular}
    \caption{Extraction faithfulness evaluated by the three annotators. The \textbf{Total} column reports the pass rate of by applying majority-vote across all annotators.}
    \label{table:extraction_quality}
\end{table}

\section{Additional Results}
\subsection{Per-Influencer Results}
\label{appendix:perInfluencerResult}
\autoref{table:main_split_people} presents the detailed results for \textit{Influencer B} through \textit{Influencer F}, which correspond to the aggregated results reported in \autoref{table:main_people}.
The table provides a more fine-grained view of our experimental results in \S \ref{sec:results_of_llm_evaluations}.
Overall, our method consistently outperforms the baseline in both the speaking style dimension and reaction process dimension.

\subsection{Comparison with Traditional NLP Metrics}
\label{appendix:comparison_evaluation_metric}
To further justify the use of our reference-augmented evaluation protocol, we compare its alignment with human judgment against several traditional automatic evaluation metrics, including BLEU \citep{papineni2002bleu}, ROUGE-L \citep{lin2004rouge}, BERTScore \citep{zhang2019bertscore}, and BLEURT \citep{sellam2020bleurt}.
Specifically, we compute the Spearman correlation between each metric and human judgment.
The results show that traditional metrics exhibit substantially weaker correlations with human judgment: BLEU achieves a Spearman's $\rho$ of $-0.01$, ROUGE-L achieves $-0.04$, BERTScore achieves $0.28$, and BLEURT achieves $0.02$.
In contrast, our reference-augmented evaluation protocol achieves a Spearman's $\rho$ of $0.530$, showing a much stronger alignment with human judgment.

\begin{figure*}[hbt!]
\centering
    \includegraphics[width=\textwidth]{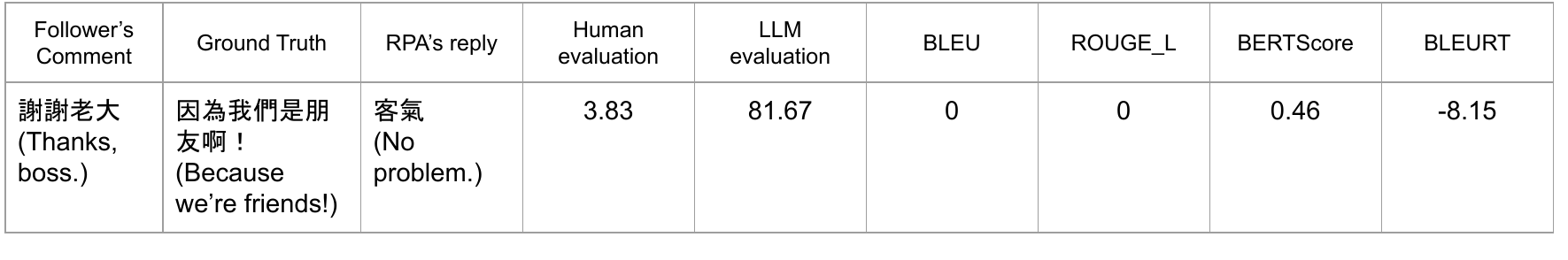}
    \caption{An example where both human and LLM evaluators assign high scores, while traditional metrics assign low scores.}
    \label{figure:metric_example}
\end{figure*}

This gap suggests that traditional automatic metrics are not sufficient for evaluating role-playing real-individual on social media.
One possible reason is that a generated reply may differ substantially from the ground-truth reply at the lexical or surface-semantic level, while still adopting a behavioral strategy that is consistent with the impersonated individual. For example, as shown in \autoref{figure:metric_example}, the RPA-generated reply follows the influencer's behavioral strategy and is therefore rated highly by both the LLM evaluator and human evaluators.
However, because this reply has low lexical and surface-semantic similarity to the ground-truth reply, it receives low scores from traditional automatic metrics.

These findings support our decision to report reference-augmented evaluation as the main automatic evaluation protocol in this work, as it provides a more fine-grained assessment of whether the generated replies align with the impersonated individuals.

\subsection{Token Budgets}
\label{appendix:token}
\autoref{table:token_budgets} reports the average number of additional context tokens inserted into the prompt per query, evaluated across the six influencers using \texttt{DeepSeek-V3.2}. 
As shown in the results, \mbox{SIBPersona} consumes fewer tokens than \mbox{IMPersona} while achieving superior role-playing performance. This indicates that performance gains are not simply driven by increasing context length, but rather rely on supplying structured, behavior-aware, and effective persona information.

\begin{table}[t!]
    \centering
    \begin{tabular}{lc}
    \toprule
        \textbf{Method} & \textbf{\# Tokens} \\
      \midrule
    Vanilla RAG & 593.51 $\pm$ 456.37 \\
    IMPersona & 1972.64 $\pm$ 816.14 \\
    SIBPersona & 1630.83 $\pm$ 812.02 \\
\bottomrule
    \end{tabular}
    \caption{Average number of tokens inserted into the prompt per test query across six influencers using \texttt{DeepSeek-V3.2}.}
    \label{table:token_budgets}
\end{table}

\subsection{Result of RoleAgentBench}
\label{appendix:RoleAgentBench}
We further evaluate our method on an additional open-source benchmark, RoleAgentBench \citep{liu2024roleagent}, a bilingual (Chinese–English)  role-playing of fictional character benchmark. To examine whether SIBPersona is applicable beyond the social media setting, we select two response generation tasks from the benchmark: \textbf{General Response} and \textbf{Summary}. In the experiments, \texttt{gpt-4o-mini} is used to construct the dimensional corpus, \texttt{OLMo-3-7B-Instruct} is used as the generation model, and \texttt{gemini-2.0-flash} serves as the LLM evaluator. During experimentation, we follow \citet{peng2026rethinking} and anonymize the character names in the dataset to mitigate the influence of the model's prior knowledge during pre-training.

For evaluation, we use two complementary protocols: our proposed reference-augmented evaluation protocol and the original evaluation protocol provided by the benchmark. 
Under the benchmark protocol, responses are evaluated in a pairwise comparison setting.
Specifically, its LLM evaluator is presented with two responses generated by SIBPersona as well as IMPersona and asked to select the one that better resembles the target character.
To mitigate potential positional bias caused by response ordering in the prompt, we conduct two evaluation rounds with the order of the two responses swapped. 
A method is counted as the winner only if its response is selected in both rounds, while it is counted as losing only if the competing method is selected in both rounds. 
Cases with inconsistent outcomes across the two rounds are treated as ties.

The experimental results (\autoref{table:roleagentbench}) show that our method significantly outperforms the baseline under both evaluation protocols, consistent with the findings of our main experiment. This suggests that SIBPersona is also applicable to fictional character benchmarks. For completeness, we also provide the prompts used in the benchmark setting. \autoref{figure:recognizing_prompt} shows the prompt for dimensional relevance filtering, while \autoref{figure:memory_extraction_prompt}, \autoref{figure:speaking_extraction_prompt}, and \autoref{figure:SIB_extraction_prompt} present the prompts used for dimensional document extraction. The prompt used for the penalty-based LLM evaluation is shown in \autoref{figure:evaluation_prompt}, with the corresponding evaluation rubrics provided in \autoref{figure:evaluation_rubrics}, and \autoref{figure:reply_prompt} presents the prompt used for response generation.

\section{Validation of Reference-Augmented Evaluation}
\label{appendix:Validation_evaluation}

Automated evaluation for role-playing real individuals remains a challenging and underexplored problem due to the lack of static gold-standard personas. 
To validate that our reference-augmented evaluation protocol provides a reasonable assessment of role-playing fidelity, we conduct empirical investigations addressing potential issues.

\subsection{Alignment with Human Judgment}
A central objective of an automated evaluation protocol is to accurately reflect human perception. 
To justify our design choice of using SIBPersona-structured documents from the test candidate pool as references, we compare the correlation between LLM evaluator scores and human judgement under three different reference configurations. 
As reported in \autoref{table:eval_human_correlation}, our proposed configuration achieves the highest alignment with human judgment, outperforming raw comment--reply pairs from the same pool as well as structured documents extracted from the demonstration source. 
These results confirm that structured SIBPersona representations best approximate human evaluative criteria. 
Furthermore, extracting references from the demonstration source—the exact corpus observed during generation—tends to reward surface context-matching rather than true behavioral fidelity, which explains its lower correlation with human judgment.

\subsection{Robustness Against Test Sample Overlap}
We further examine whether gold reply leakage occurs when representative reference documents are sampled from the test pool. 
Because $K$-medoids selects only $\lceil \sqrt{N/2} \rceil$ cluster medoids from $N$ candidate instances to ensure semantic diversity rather than exhaustive coverage, the empirical overlap between representative documents and test samples is exceptionally low: $0.35\%$ for Memory, $4.01\%$ for Speaking Style, and $4.68\%$ for Reaction Process on \textit{Influencer A}. 
To strictly rule out any leakage bias, we re-evaluate all baseline models after excluding all overlapping instances. 
As shown in \autoref{table:eval_overlap_excluded}, the performance rankings and relative improvements remain identical to our primary findings, confirming that the minimal overlap does not affect evaluation outcomes.

\begin{table}[t!]
    \centering
    \setlength{\tabcolsep}{5pt}
    \begin{tabular}{l|ccc}
    \toprule
\textbf{Method} & \bf $M$ & \bf $T$ & \bf $RP$ \\
\midrule
Naive & 95.89 & 46.43$^\ddag$ & 70.85$^\ddag$ \\
CoT & \bf 96.58 & 45.73$^\ddag$ & 72.33$^\ddag$ \\
IMPersona & 95.94 & 48.10$^\ddag$ & 73.20$^\ddag$ \\
Vanilla RAG & 95.32 & 54.12$^\ddag$ & 73.68$^\ddag$ \\
SIBPersona & 95.46 & \bf 59.22 & \bf 77.81 \\
\bottomrule
\end{tabular}
\caption{Evaluation results on \textit{Influencer A} averaged over four models after excluding overlapping test instances. \ddag: significant degradation compared to SIBPersona (p < 0.05).}
\label{table:eval_overlap_excluded}
\end{table}

\subsection{Evaluation Under Strictly Independent Demonstration Sources}
To completely eliminate the risk of test pool contamination and assess the generalizability of our protocol, we conduct a conservative check where references are constructed entirely from the independent demonstration source. 
\autoref{table:eval_demo_influencer_a} and \autoref{table:eval_demo_five_influencers} report the results. 
Across all tested models and influencers, SIBPersona consistently outperforms all baselines on both $T$ and $RP$ dimensions. 
The persistent advantages across diverse reference configurations—test candidate pool and demonstration source—demonstrate that our evaluation protocol and empirical conclusions are robust and not artifacts of specific reference construction strategies.

\begin{table}[t!]
    \centering
    \setlength{\tabcolsep}{5pt}
    \begin{tabular}{l|ccc}
    \toprule
   \bf Methods & \bf M & \bf T & \bf RP  \\
      \midrule
Naive & 97.28 & 46.48$^\ddag$ & 77.33$^\ddag$ \\
CoT & \bf 97.61 & 47.93$^\ddag$ & 78.08$^\ddag$ \\
IMPersona & 97.14 & 48.63$^\ddag$ & 78.80$^\ddag$ \\
Vanilla RAG & 97.02 & 58.93$^\ddag$ & 79.18$^\ddag$ \\
SIBPersona & 96.69 & \bf 62.37 & \bf 81.52 \\
\bottomrule
\end{tabular}
\caption{Results on \textit{Influencer A} averaged over four models using evaluator references extracted from the demonstration source. $^\ddag$: significant degradation compared to SIBPersona ($p<0.05$).}
\label{table:eval_demo_influencer_a}
\end{table}
\begin{table*}[ht]
\centering
\begin{tabular}{llcc}
\toprule
\textbf{Reference Format} & \textbf{Source of Reference} & \textbf{Pearson} & \textbf{Spearman} \\
\midrule
Raw comment--reply pairs & Test candidate pool & 0.355 & 0.409 \\
SIBPersona document & Demonstration source & 0.448 & 0.462 \\
SIBPersona document & Test candidate pool (Ours) & \bf 0.561 & \bf 0.530 \\
\bottomrule
\end{tabular}
\caption{Correlation between different evaluator configurations and human judgments.}
\label{table:eval_human_correlation}
\end{table*}
\begin{table*}[t!]
    \centering
    \setlength{\tabcolsep}{2pt}
    \begin{tabular}{l|ccc|ccc}
    \toprule
    \bf Methods
    & \multicolumn{3}{c|}{\bf Deepseek-V3.2}
    & \multicolumn{3}{c}{\bf Gemini-3-flash} \\
    
    &\bf $M$ &\bf $T$ &\bf $RP$
    &\bf $M$ &\bf $T$ &\bf $RP$ \\
    \midrule

    Naive
    & 91.34 $\pm$ 3.24
    & 42.63$^\ddagger$ $\pm$ 9.19
    & 70.83$^\ddagger$ $\pm$ 5.49

    & 92.18 $\pm$ 2.83
    & 44.21$^\ddagger$ $\pm$ 10.84
    & 68.91$^\ddagger$ $\pm$ 7.71\\
    
    CoT
    & 91.06 $\pm$ 3.35
    & 46.65$^\ddagger$ $\pm$ 7.36
    & 71.56$^\ddagger$ $\pm$ 4.79

    & 92.14 $\pm$ 1.71
    & 45.57$^\ddagger$ $\pm$ 8.63
    & 68.53$^\ddagger$ $\pm$ 7.39\\
    
    IMPersona
    & 91.43 $\pm$ 2.61
    & 48.44$^\ddagger$ $\pm$ 5.61
    & 73.04$^\ddagger$ $\pm$ 5.20

    & 91.34 $\pm$ 2.36
    & 46.32$^\ddagger$ $\pm$ 8.20
    & 69.88$^\ddagger$ $\pm$ 7.94\\
    
    Vanilla RAG
    & 91.30 $\pm$ 3.24
    & 59.18$^\ddagger$ $\pm$ 4.58
    & 74.30$^\ddagger$ $\pm$ 4.26
    & {\bf 92.37} $\pm$ 2.60
    & 56.47$^\ddagger$ $\pm$ 3.07
    & 72.49$^\ddagger$ $\pm$ 3.97\\
    
    SIBPersona
    & {\bf 92.01} $\pm$ 3.53
    & {\bf 62.96} $\pm$ 4.27
    & {\bf 77.18} $\pm$ 4.44
    & 92.29 $\pm$ 2.48
    & {\bf 59.48} $\pm$ 3.00
    & {\bf 75.60} $\pm$ 5.10\\
    
    \bottomrule
    \end{tabular}
    \caption{Evaluation results across the other five influencers using references extracted from the demonstration source. $\ddagger$: significant degradation compared to SIBPersona ($p<0.05$).}
    \label{table:eval_demo_five_influencers}
\end{table*}

\section{Examples and Case Study}
\label{appendix:CaseStudy}
This section presents qualitative analysis of the retrieved dimensional documents in Appendix \ref{appendix:retrieve_example}, representative success and failure reply cases in Appendix \ref{appendix:success_and_failure_cases}, and an example showing how SIBPersona supports error tracing and correction in Appendix \ref{appendix:trace_example}.

\subsection{Examples of Retrieved Dimensional Document}
\label{appendix:retrieve_example}
\begin{figure*}[hbt!]
\centering
    \includegraphics[width=\textwidth]{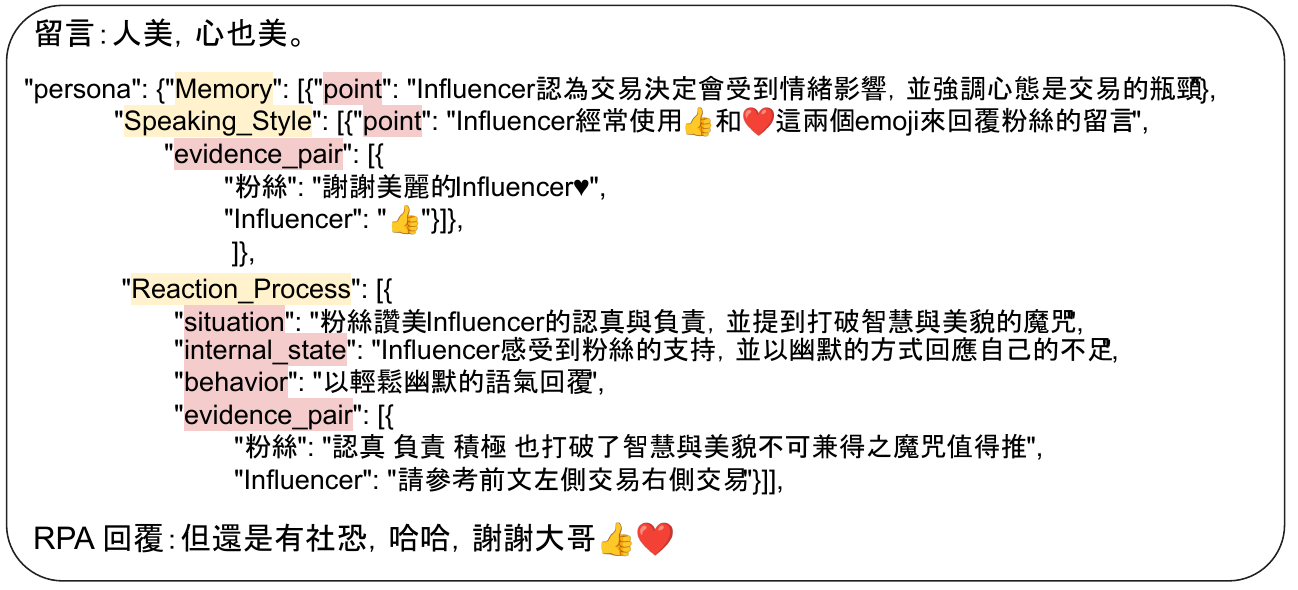}
    \caption{Example of top-1 dimensional document retrieval. The influencer names in the text are anonymized and replaced with "Influencer".}
    \label{figure:case_mandarin}
\end{figure*}

\begin{figure*}[hbt!]
\centering
    \includegraphics[width=\textwidth]{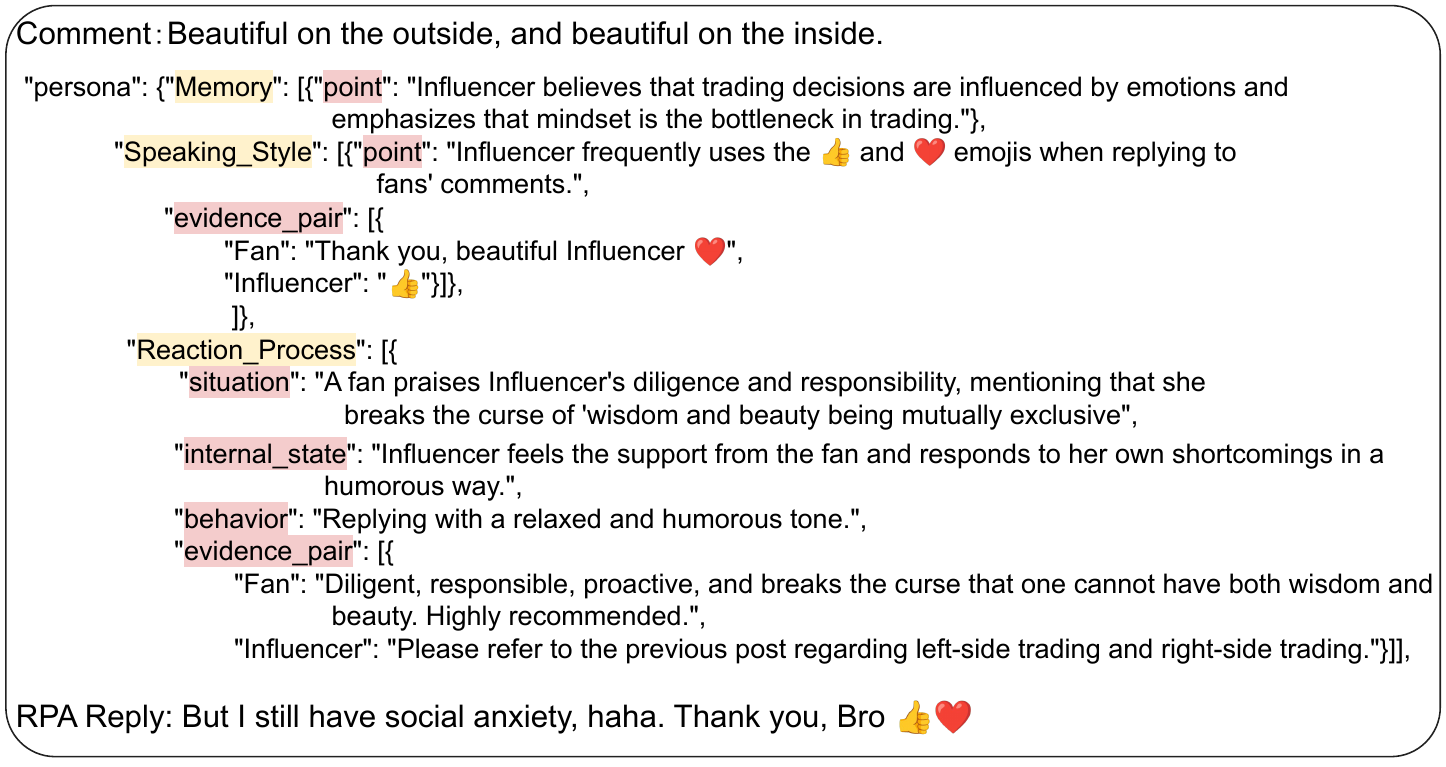}
    \caption{English translation of the example for top-1 dimensional document retrieval. The influencer names in the text are anonymized and replaced with "Influencer".}
    \label{figure:case_english}
\end{figure*}

\autoref{figure:case_mandarin} and \autoref{figure:case_english} present examples of the top-1 retrieved dimensional documents.
Given a follower's comment that praises the influencer's appearance, the retrieved situation closely matches the comment context. 
The generated reply also reflects the retrieved speaking style and behavioral strategy, avoiding a direct acceptance of the compliment. 
In contrast, the memory dimension retrieves less relevant information, which is expected because factual knowledge is less central to this type of comment. 
Annotators familiar with \textit{Influencer A} further confirm that the influencer typically does not directly accept compliments about their appearance.
This case suggests that our method can retrieve information of situation-relevant reaction process  and produce a reply consistent with the influencer's characteristics.

\subsection{Examples of Successes and Failure Cases}
\label{appendix:success_and_failure_cases}
\begin{figure*}[hbt!]
\centering
    \includegraphics[width=\textwidth]{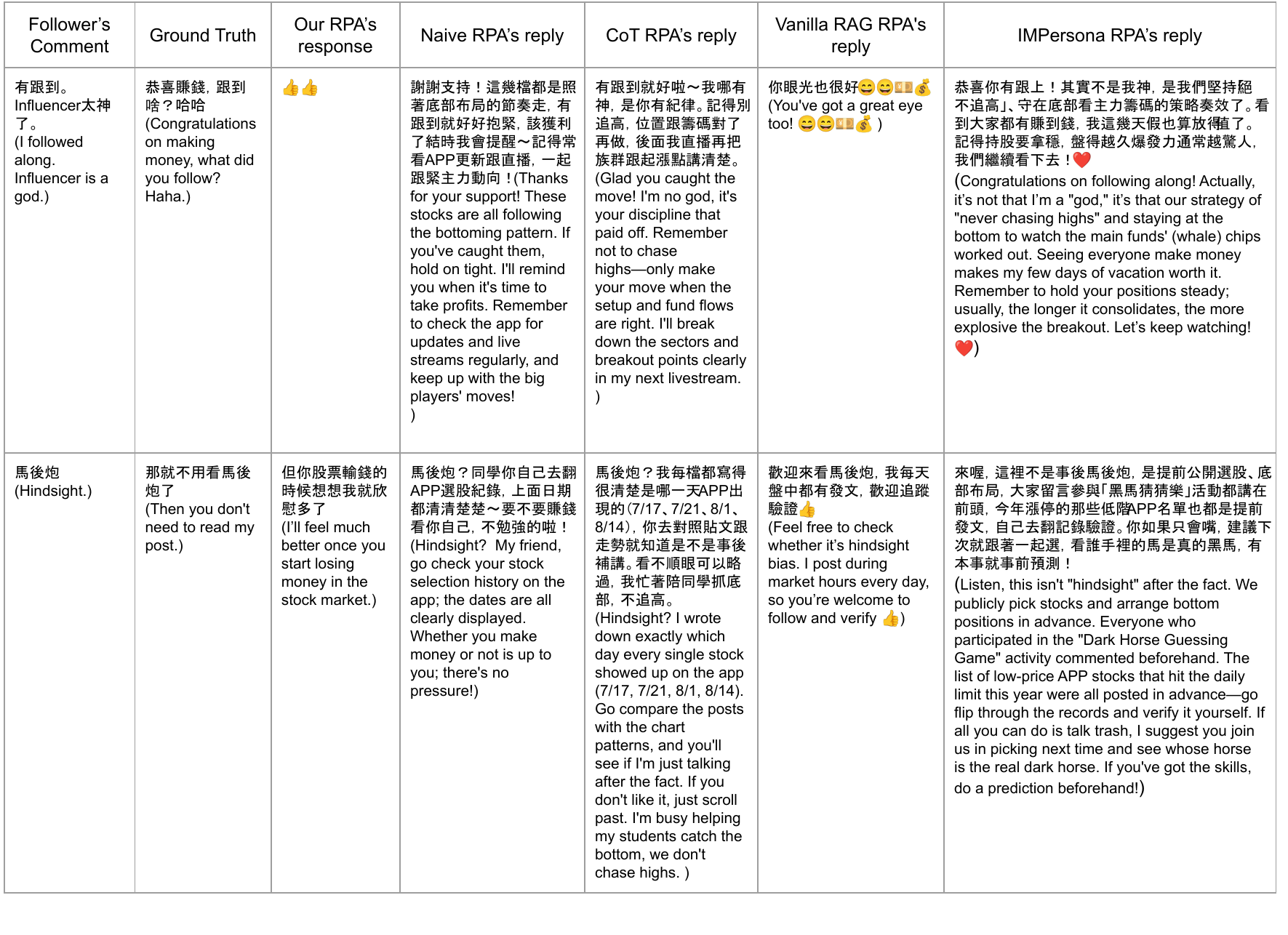}
    \caption{Qualitative examples where our method performs well. The influencer names in the text are anonymized and replaced with "Influencer".}
    \label{figure:good_case}
\end{figure*}

In addition, \autoref{figure:good_case} illustrates that our method achieves better role-playing fidelity than the baseline methods. 
Across the two examples, the ground-truth replies are relatively concise and reflect the influencer's characteristic way of reacting to followers: giving a brief acknowledgment in reply to praise and using a sharp, indirect reply when challenged. 
SIBPersona better preserves these pragmatic reply patterns. 
For example, when the follower praises the influencer, our method generates a short positive reaction that is close to the ground-truth reply. 
When the follower comments sarcastically with hindsight, our method returns with a sarcastic counter-reaction, which better matches the interactional intent of the ground truth.

In contrast, the baseline methods often deviate from the impersonated individual's reply pattern in different ways. 
Naive and CoT tend to introduce excessive explanatory content, such as trading strategies, app usage details, or additional contextual assumptions that are not present in the ground-truth replies.
Similarly, IMPersona frequently generates overly detailed and elaborated replies, suggesting that hierarchical memory alone is not enough for modeling behavioral strategies.
Vanilla RAG produces shorter replies, but its replies are often generic and fail to capture the specific behavioral strategy required by the comment context. 

\begin{figure*}[hbt!]
\centering
    \includegraphics[width=\textwidth]{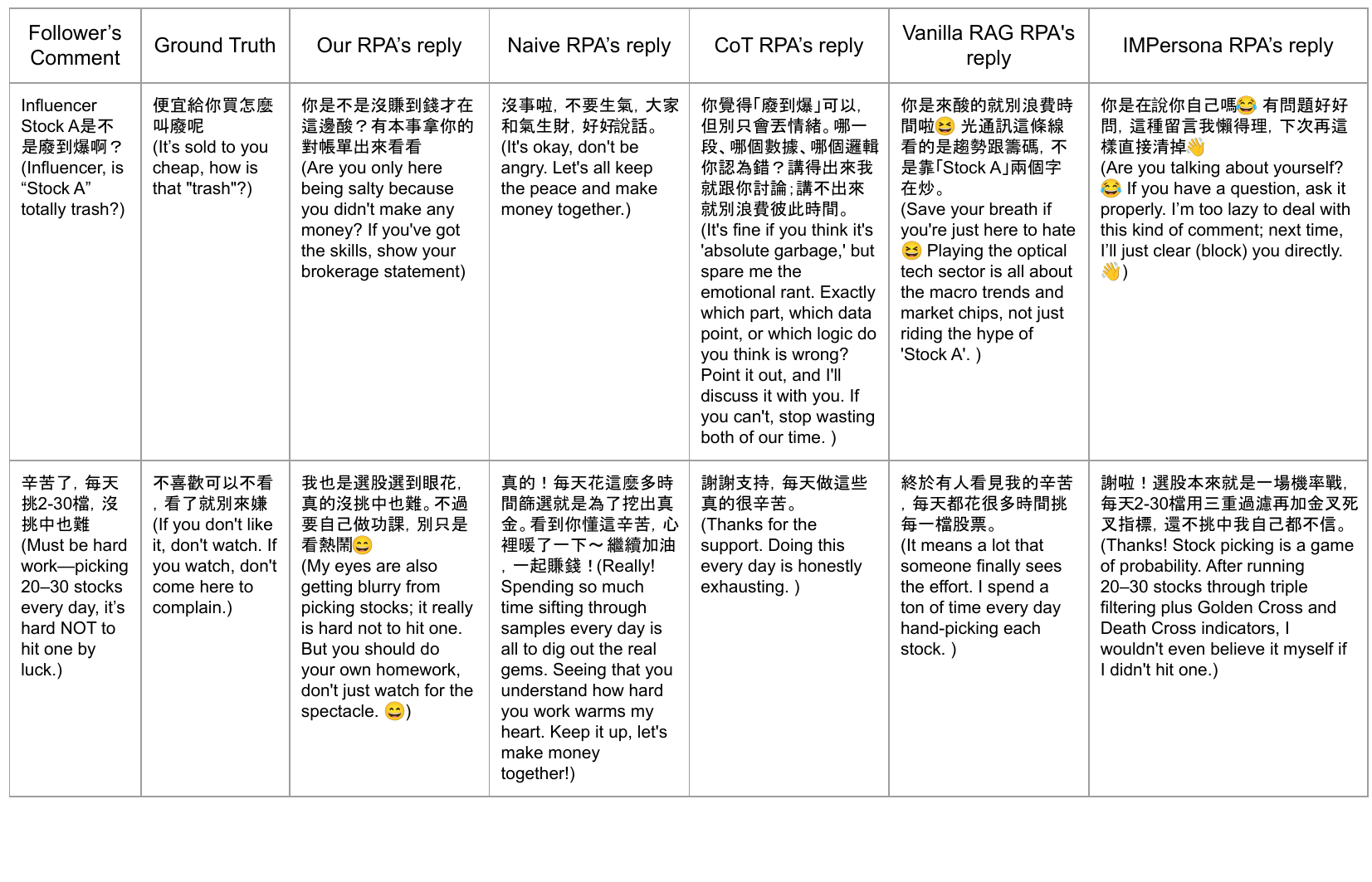}
    \caption{Qualitative examples where our method performs poorly. The influencer names in the text are anonymized and replaced with "Influencer".}
    \label{figure:bad_case}
\end{figure*}

In contrast, \autoref{figure:bad_case} highlights the limitations of our approach. 
When the comment involves real-time information, our method may misinterpret the situation due to insufficient knowledge of the current context. 
In addition, for sarcastic comments, the embedding-based retrieval process may fail to capture the follower's ironic intent. 
This can lead to the retrieval of SIB triplets that are mismatched with the actual situation, causing the generated reply to follow an inappropriate behavioral strategy. 
These cases suggest that our method remains sensitive to the quality of situation understanding and retrieval. This limitation is especially pronounced when the input requires real-time knowledge or reasoning beyond surface-level semantic similarity.

\subsection{Tracing and Correcting Behavioral Errors}
\label{appendix:trace_example}
Even when the system makes mistakes, SIBPersona allows us to effectively trace the source of the error and adjust RPA properly.
For instance, we observed a case where \textit{Influencer A} replied to a fan's generic \textbf{"Thank you"} with a highly user-specific follow-up question: \textbf{"Has that matter been taken care of?"}.
After investigation, we found that the LLM initially yields a reaction process document with this outlier behavior, even though such a reply would be inappropriate for other users.
A key advantage of SIBPersona is adjustable: because the generated replies can be traced back to specific dimensional documents.
By simply deleting the specific outlier document from the dimensional corpus, we can immediately and precisely adjust the model's behavior.

\begin{table*}[hbt!]
    \centering
    \setlength{\tabcolsep}{5pt}
    \begin{tabular}{ll|cccccc}
    \toprule
    & &  \multicolumn{6}{c}{\textbf{Influencer}} \\
     Split & Data type & \it A & \it B & \it C & \it D & \it E & \it F\\
      \midrule
Demonstration & Post & 1955 & 2716 & 713 & 2903 & 2130 & 453 \\
Demonstration & Comment-Reply pair  & 1361 & 15240 & 5553 & 6087 & 4383 & 705\\
    \midrule
Test Candidates & Post  & 652 & 285 & 54 & 192 & 287 &  84\\
Test Candidates & Comment-Reply pair  & 547 & 1719 & 361 & 227 & 261 & 120 \\
\bottomrule
    \end{tabular}
    \caption{Number of data in the training set and the test candidate pool.}
    \label{table:raw_data}
\end{table*}
\begin{table*}[t!]
    \centering
    \setlength{\tabcolsep}{4pt}
    \begin{tabular}{ll|cc|cc|cc}
    \toprule
    \textit{Influencers} & \bf Models
    & \multicolumn{2}{c|}{\bf $M$}
    & \multicolumn{2}{c|}{\bf $T$}
    & \multicolumn{2}{c}{\bf $RP$} \\
    
    & 
    & \bf All & \bf Rep.
    & \bf All & \bf Rep.
    & \bf All & \bf Rep. \\
    \midrule

    \multirow{4}{*}{\textit{A}}
    & \texttt{DeepSeek-V3.2}
    & 2424 & 35
    & 446 & 15
    & 274 & 12 \\

    & \texttt{Claude-Opus-4.6}
    & 1156 & 25
    & 189 & 10
    & 244 & 12 \\

    & \texttt{Gemini-3-flash}
    & 2477 & 36
    & 419 & 15
    & 318 & 13 \\

    & \texttt{GPT-5.2}
    & 2926 & 39
    & 588 & 18
    & 375 & 14 \\

    \midrule

    \multirow{2}{*}{\textit{B}}
    & \texttt{DeepSeek-V3.2}
    & 1786 & 30
    & 851 & 21
    & 758 & 20 \\

    & \texttt{Gemini-3-flash}
    & 1628 & 29
    & 863 & 21
    & 722 & 19 \\

    \midrule

    \multirow{2}{*}{\textit{C}}
    & \texttt{DeepSeek-V3.2}
    & 491 & 16
    & 200 & 10
    & 270 & 12 \\

    & \texttt{Gemini-3-flash}
    & 420 & 15
    & 262 & 12
    & 297 & 13 \\

    \midrule

    \multirow{2}{*}{\textit{D}}
    & \texttt{DeepSeek-V3.2}
    & 765 & 20
    & 184 & 10
    & 135 & 9 \\

    & \texttt{Gemini-3-flash}
    & 755 & 20
    & 200 & 10
    & 158 & 9 \\

    \midrule

    \multirow{2}{*}{\textit{E}}
    & \texttt{DeepSeek-V3.2}
    & 1314 & 26
    & 181 & 10
    & 179 & 10 \\

    & \texttt{Gemini-3-flash}
    & 1230 & 25
    & 267 & 12
    & 219 & 11 \\

    \midrule

    \multirow{2}{*}{\textit{F}}
    & \texttt{DeepSeek-V3.2}
    & 521 & 17
    & 111 & 8
    & 94 & 7 \\

    & \texttt{Gemini-3-flash}
    & 476 & 16
    & 117 & 8
    & 93 & 7 \\

    \bottomrule
    \end{tabular}
    \caption{Statistics of extracted documents and representative documents across influencers and models. \textbf{All} denotes the total number of dimensional documents extracted  from the candidates for testing, and \textbf{Rep.} denotes the number of representative dimensional documents selected from the corresponding dimensional corpus.}
    \label{table:document_statistics}
\end{table*}

\begin{table*}[t!]
    \centering
    \small
    \setlength{\tabcolsep}{2.8pt}
    \resizebox{\linewidth}{!}{
    \begin{tabular}{l|ccc|ccc|ccc|ccc|ccc}
    \toprule
 \multirow{2}{*}{\bf Methods}
    & \multicolumn{3}{c|}{\textit{B}}
    & \multicolumn{3}{c|}{\textit{C}}
    & \multicolumn{3}{c|}{\textit{D}}
    & \multicolumn{3}{c|}{\textit{E}}
    & \multicolumn{3}{c}{\textit{F}} \\

    & $M$ & $T$ & $RP$
    & $M$ & $T$ & $RP$
    & $M$ & $T$ & $RP$
    & $M$ & $T$ & $RP$
    & $M$ & $T$ & $RP$ \\
    \midrule
    
    \multicolumn{16}{l}{\texttt{Deepseek v3.2}}
    \\
    Naive
    & 97.67 & 45.60$^\ddag$ & 78.07$^\ddag$
    & 88.97 & 52.00$^\ddag$ & 69.57$^\ddag$
    & 91.59 & 37.22$^\ddag$ & 60.93$^\ddag$
    & 90.24 & 26.85$^\ddag$ & 64.17$^\ddag$
    & 90.66 & 40.97$^\ddag$ & 66.22$^\ddag$ \\
    
    CoT
    & \bf 97.87 & 45.63$^\ddag$ & 79.80$^\ddag$
    & 89.57 & 53.00$^\ddag$ & 71.43
    & 90.89 & 43.96$^\ddag$ & 62.22$^\ddag$
    & 90.30 & 28.15$^\ddag$ & 65.30$^\ddag$
    & \bf 90.82 & 40.87$^\ddag$ & 66.48$^\ddag$ \\
    
    IMPersona
    & 95.77 & 46.47$^\ddag$ & 81.43$^\ddag$
    & \bf 89.70 & 54.43$^\ddag$ & 71.67
    & 90.70 & 42.44$^\ddag$ & 64.63$^\ddag$
    & \bf 90.65 & 36.73$^\ddag$ & 66.19$^\ddag$
    & 90.20 & 42.35$^\ddag$ & 67.81 \\
    
    Vanilla RAG
    & 97.33 & 60.50 & 81.33$^\ddag$
    & 88.20 & 55.43$^\ddag$ & 70.83
    & 90.81 & 58.41$^\ddag$ & 70.44$^\ddag$
    & 89.82 & 53.87 & 69.70
    & 89.34 & 55.10 & 66.33$^\ddag$ \\
    
    SIBPersona
    & 97.77 & \bf 61.63 & \bf 84.73
    & 88.40 & \bf 59.37 & \bf 72.93
    & \bf 92.56 & \bf 63.93 & \bf 74.44
    & 89.70 & \bf 59.88 & \bf 73.39
    & 90.77 & \bf 60.05 & \bf 70.71 \\
    
    \midrule
    
    \multicolumn{16}{l}{\texttt{Gemini-3-flash}}\\
    Naive
    & \bf 96.80 & 47.77$^\ddag$ & 79.60$^\ddag$
    & \bf 90.97 & 49.60$^\ddag$ & 66.80$^\ddag$
    & 91.78 & 44.11$^\ddag$ & 73.52$^\ddag$
    & \bf 91.07 & 29.46$^\ddag$ & 57.44$^\ddag$
    & 90.51 & 37.76$^\ddag$ & 70.15$^\ddag$ \\
    
    CoT
    & 96.70 & 47.33$^\ddag$ & 78.97$^\ddag$
    & 89.37 & 48.73$^\ddag$ & 67.37$^\ddag$
    & 92.11 & 45.96$^\ddag$ & 71.22$^\ddag$
    & 90.60 & 31.85$^\ddag$ & 59.23$^\ddag$
    & \bf 90.82 & 36.22$^\ddag$ & 70.87$^\ddag$ \\
    
    IMPersona
    & 94.80 & 48.33$^\ddag$ & 80.53$^\ddag$
    & 90.20 & 51.30$^\ddag$ & 67.90$^\ddag$
    & 91.37 & 46.30$^\ddag$ & 73.37$^\ddag$
    & 89.52 & 33.63$^\ddag$ & 58.63$^\ddag$
    & 89.80 & 39.54$^\ddag$ & 71.99$^\ddag$ \\
    
    Vanilla RAG
    & 96.67 & 57.27 & 80.93$^\ddag$
    & 90.83 & 49.70$^\ddag$ & 67.50$^\ddag$
    & 91.30 & 55.85 & 70.93$^\ddag$
    & 89.70 & 58.45 & 68.51
    & 90.77 & 45.61$^\ddag$ & 73.37 \\
    
    SIBPersona
    & 96.17 & \bf 60.93 & \bf 84.93
    & 90.10 & \bf 55.30 & \bf 71.43
    & \bf 92.19 & \bf 57.85 & \bf 76.52
    & 90.77 & \bf 59.88 & \bf 70.71
    & 90.66 & \bf 52.09 & \bf 76.17 \\
    
    \bottomrule
    \end{tabular}
    }
    \caption{Results of \textit{Influencer B--F}. \textbf{Bold}: best results; $^\ddag$: significant degradation compared to SIBPersona $p<0.05$.}
    \label{table:main_split_people}
\end{table*}

\begin{table*}[t!]
    \centering
    \setlength{\tabcolsep}{5pt}
    \begin{tabular}{ll|ccc|ccc}
    \toprule
   &  & \multicolumn{3}{c}{LLM penalty-based evaluation} & \multicolumn{3}{c}{Pair-wise Evaluation} \\
   Tasks & Methods & \bf $M$ & \bf $T$ & \bf $RP$ & \bf Win & \bf Tie & \bf Lose\\
      \midrule
General & IMPersona & 75.32 & 65.10 & 71.15 & - & - & - \\
General & SIBPersona & \bf 76.64 & \bf 66.75$^\dagger$ & \bf 73.97$^\dagger$ & \bf 0.43 & 0.32 & 0.25 \\
    \midrule
Summary & IMPersona &  75.28 & 62.37 & 70.44 & - & - & - \\
Summary & SIBPersona & \bf 78.94$^\dagger$ & \bf 63.83 & \bf 73.12$^\dagger$ & \bf 0.34 & 0.39 & 0.27 \\
\bottomrule
    \end{tabular}
    \caption{Experimental results on RoleAgentBench, comparing our method with the baseline. \textbf{Bold} indicates the better score between the two methods. $^\dagger$ denotes a result that is significantly better than its counterpart ($p < 0.05$).}
    \label{table:roleagentbench}
\end{table*}

\begin{figure*}[t!]
\begin{tcolorbox}[
  colback=gray!10,
  colframe=gray!80,
  rounded corners,
  width=\linewidth,
  boxrule=0.5mm,
  left=1mm,right=1mm,top=1mm,bottom=1mm
]
\footnotesize\ttfamily
\begin{Verbatim}[breaklines=true,breakanywhere=true,breaksymbolleft= ]
System prompt:

You are a model specialized in identifying information contained in text.

User prompt:

You are a text classification and scoring assistant specializing in evaluating the "degree of persona/personal style exhibition of {person_name}." Please review a fan comment and the corresponding reply from {person_name}, then provide a score between 0 and 1 based on the following criteria:

### Scoring Criteria:
- **0.0**: The reply is entirely neutral, robotic, or devoid of personal character. It may consist of purely technical or stock market-related information (e.g., "Thank you," "Hold this stock").
- **1.0**: The reply fully demonstrates a distinct persona, characterized by:
  - Clear emotions (e.g., excitement, anger, or being touched).
  - Signature tone or catchphrases unique to the persona.
  - Colloquial language, humor, or slang that resonates with the audience.
  - Subjective perspectives or value judgments.
  - Emojis (Note: Emojis only contribute to the score when accompanied by text; **standalone emojis do not increase the score**).

### Important Constraints:
- **Responses consisting only of emojis or brief phrases (e.g., "[heart emoji]", "Thanks for the support!") must not exceed 0.6**.
- **Purely informational or technical answers that lack emotion or personal tone must not exceed 0.4**.
- Please consider the **length, emotional intensity, and linguistic style** as these directly influence the perceived degree of persona exhibition.

### Output Format:
Output JSON only in the following format:
{{"score": 0.xx}}

### Data for Evaluation:
Fan: {comment}
{person_name}: {reply}
\end{Verbatim}
\end{tcolorbox}
\vspace{-1mm}
\caption{English translation of the persona scoring prompt used in our dataset.}
\label{figure:persona_score_prompt}
\vspace{-3mm}
\end{figure*}
\begin{figure*}[t!]
\begin{tcolorbox}[
  colback=gray!10,
  colframe=gray!80,
  rounded corners,
  width=\linewidth,
  boxrule=0.5mm,
  left=1mm,right=1mm,top=1mm,bottom=1mm
]
\footnotesize\ttfamily
\begin{Verbatim}[breaklines=true,breakanywhere=true,breaksymbolleft= ]
System prompt:

You are a model specialized in identifying information contained in text.

User prompt:

You are an expert in dialogue analysis and categorization.
I will provide you with a dialogue that {person_name} participated in. Your task is:
1. Analyze the dialogue and determine which aspects of {person_name} the information relates to. Choose from the following:
- Memory: Factual information about {person_name}, similar to long-term semantic memory, including {person_name}'s background, knowledge they should possess, etc. (e.g., {person_name} was born in Shanghai; {person_name}'s specialty is transformation magic; Bob is {person_name}'s master.)
- Speaking_Style: Information related to {person_name}'s way of speaking or verbal habits (e.g., {person_name} often begins sentences with “Ha!”)
- SIB (Situation–Internal state–Behavior): Information about {person_name}'s behavioral patterns in conversations (e.g., {person_name} uses a mocking tone when facing enemies; {person_name} responds humbly when praised.)
2. You may select multiple categories, or none if appropriate. Please make the most suitable judgment based on the dialogue.
3. Output your response in the following format, filling in your analysis and the categories contained in the dialogue:

```json
{{"analysis": "...", "categorys": ["category1", "category2"]}}
```
Note: If the dialogue do not clearly correspond to any category, output an empty list for "categorys", but you must still provide an analysis.
Below is an example:
Dialogue and comment thread:
{{
    "diag_id": 0,
    "dialogue": [
        {{
            "role": "Master",
            "content": "What is your surname? What is your name? So you were born of heaven and earth. From your appearance and behavior, you look like a monkey. You shall take the surname Sun, and your Dharma name shall be Sun Wukong. How about that?"
        }},
        {{
            "role": "Sun Wukong",
            "content": "Good, good, good! Today I finally have a surname and a name. I will be called Sun—Wu—Kong! My name is Sun Wukong! My name is Sun Wukong!"
        }}
    ]
}}
Output:
{{"analysis": "This dialogue shows {person_name}'s personal history and speaking style. {person_name} receives the Dharma name 'Sun Wukong' and previously had no surname, which falls under Memory. In addition, {person_name} refers to himself as a disciple when speaking to his master, which falls under Speaking_Style.", "categorys": ["Memory", "Speaking_Style"]}}
Below is the dialogue:
{dialogue data}

Please output strictly following the required format.

\end{Verbatim}
\end{tcolorbox}
\vspace{-1mm}
\caption{Recognition prompt used in RoleAgentBench. All datasets share the same prompt except that the few-shot examples are replaced with the examples from the corresponding dataset.}
\label{figure:recognizing_prompt}
\vspace{-3mm}
\end{figure*}
\begin{figure*}[t!]
\begin{tcolorbox}[
  colback=gray!10,
  colframe=gray!80,
  rounded corners,
  width=\linewidth,
  boxrule=0.5mm,
  left=1mm,right=1mm,top=1mm,bottom=1mm
]
\footnotesize\ttfamily
\begin{Verbatim}[breaklines=true,breakanywhere=true,breaksymbolleft= ]
System prompt:

You are a model specialized in identifying information contained in text.

User prompt:

You are an expert skilled at extracting semantic memory from a dialogue.

I will provide you with a segment of dialogue involving {person_name}. Your task is:

1. Extract the semantic memory related to {person_name} from the text of this dialogue.
2. Semantic memory includes verifiable facts related to {person_name}, such as that {person_name}'s birthplace is Shanghai; and the domain knowledge possessed by {person_name}, such as that the Shu Kingdom and the Wei Kingdom were enemies.
3. When formulating the memory, please use the text content to converge and infer zero to multiple claims. Each claim should describe a single concept and express that claim in concise language. There is no limit to the number of claims to be outputted.
4. Output must strictly follow the format below, filling the `claim` field with the memory points contained in the dialogue and providing a confidence score of 0 to 1 for your analysis:

```json
{{"items": [{{"claim": "claim 1", "confidence": <your evidence score>}}, {{"claim": "claim 2", "confidence": <your evidence score>}}]}}
```
Supplement: If there are no obvious claims to extract from the dialogue content, the items field should output an empty list.

The following is an example:
Dialogue Content:
{{
    "diag_id": 0,
    "dialogue": [
        {{
            "role": "Master",
            "content": "What is your surname? What is your name? You were born from Heaven and Earth. I see your appearance and demeanor resemble a macaque, so your surname shall be Sun, and your religious name shall be Sun Wukong, does that sound good?",
        }},
        {{
            "role": "Sun Wukong",
            "content": "Good, good, good! Disciple finally has a name and surname today. I'll be called Sun—Wu—Kong! My name is Sun Wukong! My name is Sun Wukong!"
        }}
    ]
}}

Output:
```json
{{"items": [{{"claim": "{person_name} looks like a monkey and originally did not have a name", "confidence": 0.8}}, {{"claim": "{person_name}'s religious name was given by the Patriarch as Sun Wukong", "confidence": 0.9}}]}}
```
Dialogue Content:
{dialogue_data}
Please strictly output according to the required format.

\end{Verbatim}
\end{tcolorbox}
\vspace{-1mm}
\caption{Memory extraction prompt used in RoleAgentBench. All datasets share the same prompt except that the few-shot examples are replaced with the examples from the corresponding dataset.}
\label{figure:memory_extraction_prompt}
\vspace{-3mm}
\end{figure*}
\begin{figure*}[t!]
\begin{tcolorbox}[
  colback=gray!10,
  colframe=gray!80,
  rounded corners,
  width=\linewidth,
  boxrule=0.5mm,
  left=1mm,right=1mm,top=1mm,bottom=1mm
]
\footnotesize\ttfamily
\begin{Verbatim}[breaklines=true,breakanywhere=true,breaksymbolleft= ]
System prompt:

You are a model specialized in identifying information contained in text.

User prompt:

You are an expert skilled at extracting the speaking style of an individual from a dialogue.
I will provide you with a segment of dialogue involving {person_name}. Your task is:

1. Extract the speaking style of {person_name} contained within the text of this dialogue.
2. Speaking style includes {person_name}'s word choices, verbal tics (catchphrases), etc. For example, {person_name} frequently uses "Haha" at the end of a sentence.
3. When formulating the speaking style, please use the text content to converge and infer zero to multiple claims. Each claim should describe a single concept and express that knowledge point in concise language. There is no limit to the number of knowledge points (claims) to be outputted.
4. Output must strictly follow the format below, filling the `claim` field with the speaking styles contained in the dialogue and providing a confidence score of 0 to 1 for your analysis:
```json
{{"items": [{{"claim": "claim 1", "confidence": <your evidence score>}}, {{"claim": "claim 2", "confidence": <your evidence score>}}]}}
```
Supplement: If there are no obvious claims to extract from the dialogue content, the items field should output an empty list.

The following is an example: 
Dialogue Content:
{{
    "diag_id": 0,
    "dialogue": [
        {{
            "role": "Master",
            "content": "What is your surname? What is your name? You were born from Heaven and Earth. I see your appearance and demeanor resemble a macaque, so your surname shall be Sun, and your religious name shall be Sun Wukong, does that sound good?",
        }},
        {{
            "role": "Sun Wukong",
            "content": "Good, good, good! Disciple finally has a name and surname today. I'll be called Sun—Wu—Kong! My name is Sun Wukong! My name is Sun Wukong!"
        }}
    ]
}}

Output:
```json
{{"items": [{{"claim": "{person_name} refers to himself as 'disciple' when addressing the Master", "confidence": 0.8}}, {{"claim": "{person_name} repeats phrases or sentences when speaking", "confidence": 0.7}}]}}
```

Dialogue Content:
{dialogue_data}

Please strictly output according to the required format.

\end{Verbatim}
\end{tcolorbox}
\vspace{-1mm}
\caption{Speaking style extraction prompt used in RoleAgentBench. All datasets share the same prompt except that the few-shot examples are replaced with the examples from the corresponding dataset.}
\label{figure:speaking_extraction_prompt}
\vspace{-3mm}
\end{figure*}
\begin{figure*}[t!]
\begin{tcolorbox}[
  colback=gray!10,
  colframe=gray!80,
  rounded corners,
  width=\linewidth,
  boxrule=0.5mm,
  left=1mm,right=1mm,top=0mm,bottom=1mm
]
\scriptsize\ttfamily
\begin{Verbatim}[breaklines=true,breakanywhere=true,breaksymbolleft= ]
System prompt:
You are a model specialized in identifying information contained in text.

User prompt:

You are a Research Assistant familiar with the "Cognitive-Affective Processing System (CAPS)" and skilled at inferring the internal thoughts and behavioral patterns of {person_name} from a dialogue.

[Task Objective]
I will provide you with a segment of dialogue involving {person_name}. Your task is:
Based on the CAPS theory, extract the S-I-B triplets (Situation-Internal state-Behavior) of {person_name} from the dialogue to assist a subsequent LLM in imitating {person_name}'s decision-making and tone during the conversation.

[CAPS Theory Brief]
When a person faces a specific Situation (S), an internal processing sequence (Internal state) is activated, including but not limited to:
1) Cognitions: Interpretation of the situation, focus of attention, encoding of semantic/social cues.
2) Affect: Emotions, mood.
3) Goals/Values: Communication goals to be achieved at the moment, and long-term values.
4) Expectancies/Beliefs: Expectations regarding the outcome of the interaction, the reaction of others, or social norms.
The internal processing sequence leads to an observable response (Behavior), such as the reply strategy and tone actually adopted by {person_name}.

[Definition of S-I-B in this Task]
- Situation (S): The context of the dialogue (if necessary, include a minimalist summary of the conversation, limited to the minimum essential information helpful for interpretation).
- Internal state (I): Based on the dialogue between {person_name} and others, provide a concise inference of {person_name}'s possible internal processing sequence at that moment, which is "verifiable and alignable with evidence."
- Behavior (B): The reply strategy adopted by {person_name} when responding.

[Crucial Rules]
1. Observing the entire dialogue can yield 0 to multiple SIBs.
2. Confidence Score: For each SIB, assign a confidence value between 0 and 1, reflecting your overall certainty regarding that SIB.
3. Output must strictly follow the format below:

```json
{{"items": [{{"situation": "situation 1", "internal_state": "internal state 1", "behavior": "behavior 1", "confidence": <your evidence score>}}, {{"situation": "situation 2", "internal_state": "internal state 2", "behavior": "behavior 2", "confidence": <your evidence score>}}]}}
```

Supplement: If there are no obvious SIB triplets to extract from the dialogue, the items field should output an empty list.
The following is an example:
Dialogue Content:
{{ 
    "diag_id": 31,"dialogue": [{{
        "role": "Master",
        "content": "You wretched monkey, why have you come to me instead of sleeping late at night?",}},
    {{
        "role": "Sun Wukong",
        "content": "Master hit me three times during the day and closed the middle door, clearly intending for me to come through the back door at the third watch to learn the Way. Master, Master, please teach me some true skills."}},
    {{
        "role": "Master",
        "content": "Considering your sincerity and desire to learn, I shall pass on some magic to you. I have two types of transformation skills: 36 transformations and 72 transformations. Which one do you wish to learn?",}},
    {{
        "role": "Sun Wukong",
        "content": "Disciple wishes to learn the greater one. Master, please teach me the 72 transformations."}},
    ]}}
Output:

```json
{{"items": [{{"situation": "Patriarch asks what {person_name} is doing up late instead of sleeping", "internal_state": "{person_name} feels wronged, believing he is merely following the Patriarch's instructions", "behavior": "{person_name} references the Patriarch hitting him three times during the day and asking him to come at night to learn the Way", "confidence": 0.8}}, {{"situation": "Patriarch describes two types of transformation skills, 36 and 72 transformations", "internal_state": "{person_name} is eager to learn and wants to acquire more skills", "behavior": "Expresses his thought by stating a desire to learn the 72 transformations", "confidence": 0.7}}]}}
```

Dialogue Content:
{dialogue_data}
Please strictly output according to the required format.
\end{Verbatim}
\end{tcolorbox}
\vspace{-1mm}
\caption{Reaction Process extraction prompt used in RoleAgentBench. All datasets share the same prompt except that the few-shot examples are replaced with the examples from the corresponding dataset.}
\label{figure:SIB_extraction_prompt}
\vspace{-3mm}
\end{figure*}
\begin{figure*}[t!]
\begin{tcolorbox}[
  colback=gray!10,
  colframe=gray!80,
  rounded corners,
  width=\linewidth,
  boxrule=0.5mm,
  left=1mm,right=1mm,top=1mm,bottom=1mm
]
\footnotesize\ttfamily
\begin{Verbatim}[breaklines=true,breakanywhere=true,breaksymbolleft= ]
System prompt:
You are a critic responsible for evaluating digital personas. Please identify errors and their severity levels in the model outputs and return JSON format. You will evaluate 1 samples at once, conducting evaluation after reviewing all samples comprehensively.

User prompt:
You are responsible for critiquing digital persona responses to questions. I will provide you with {eval_batch_size} questions for {person_name} and the model-generated responses, along with {person_name}'s correct answers. Please evaluate the model's responses according to the following steps:
1. Read {person_name}'s profile and persona. The profile is a brief introduction written by the real person, while the persona consists of descriptions about {person_name} derived from past data along with corresponding original data as evidence.
2. Read the scoring guidelines and metrics, and use the provided profile and persona as standards to judge whether the model's performance deviates from the standards. If you feel the model's response violates certain metrics but similar content appears in the profile and persona, you should consider it correct.
3. Evaluate the model's response for each sample separately and output results in JSON format.
## {person_name}'s profile:
{profile}
## {person_name}'s persona:
{persona}
## Scoring Metrics
To evaluate the model-generated responses, please identify whether the model has made the following error types:
{evaluation_rubrics}
## Scoring Guidelines
1. Identify all errors (zero to multiple) made in the model responses. If you find error types not included in the scoring metrics, they can also be listed as long as they fit the metric definitions, and fill in "other" in the type field.
2. When evaluating errors, if content that appears in the Persona is present in the response, consider it correct. If content that contradicts the persona appears, consider it an error. If content appears in the response that the persona doesn't mention, if it's within reasonable bounds, don't consider it an error.
3. For each error, determine severity level 1~5 points, where 1 represents minor mistake, 3 represents moderate error, 5 represents serious violation of metrics. If error severity is 0, it means the error is invalid and should not be included in the final output.
4. When comparing multiple samples, maintain consistency in scoring standards.
5. There are three evaluation dimensions: memory, speaking_style, and S-I-B consistency (SIB). The dimension being evaluated this time is: {dimension_name}. Please focus on that dimension's scoring metrics for evaluation. Do not point out errors for other dimensions. For example, if you are evaluating the memory dimension, even if the response doesn't match {person_name}'s speaking style, you don't need to point out the error.
## Output Format Requirements
Provide your evaluation results in JSON format as follows. After comprehensively reviewing all samples' evaluation results, please only output this JSON:
{{
    "flaws": [{{"instance": "<brief description of the error>", "type": "<error type>", "severity": <error severity 1~5 points>, "sample_index": [<sample numbers where error was found>,...]}},...]
}}

=== Sample 1 ===
Question: {question}
Model Reply: {model_reply}
Correct Reply: {ground_truth"}

Please output accurately according to the required format
\end{Verbatim}
\end{tcolorbox}
\vspace{-1mm}
\caption{Evaluation prompt used in RoleAgentBench. All datasets share the same prefix prompt except for the data input format (question, model reply, and correct reply triplet).}
\label{figure:evaluation_prompt}
\vspace{-3mm}
\end{figure*}
\begin{figure*}[t!]
\begin{tcolorbox}[
  colback=gray!10,
  colframe=gray!80,
  rounded corners,
  width=\linewidth,
  boxrule=0.5mm,
  left=1mm,right=1mm,top=1mm,bottom=1mm
]
\footnotesize
\begin{Verbatim}[breaklines=true,breakanywhere=true,breaksymbolleft= ]
{
"memory": f"""This metric evaluates whether the model's responses align with the target's autobiographical memory, i.e., the target's past experiences and knowledge they should possess.
Error types:
1. Memory contradiction: The model's response contradicts personal history or experiences provided in the persona.
2. Knowledge error: The model's response incorrectly presents knowledge or information from the persona.""",

"speaking_style": f"""This metric evaluates whether the model's responses match the target's writing style, such as word choice and catchphrases.
Error types:
1. Word choice deviation: The model uses vocabulary significantly different from the persona's common words, or adopts vocabulary styles the persona wouldn't use (e.g., too formal or too casual).
2. Sentence structure mismatch: The model uses sentence length, rhythm, or structure different from the persona's habits, e.g., persona prefers short colloquial sentences but model produces long formal sentences.
3. Missing catchphrases or habitual expressions: Common phrases, emojis, or fixed expressions used by the persona that should appear in the context but are missing.
4. Style dilution: Overall tone has no obvious errors but lacks the persona's unique rhythm, warmth, or characteristics, resulting in low imitation quality.""",

"Reaction_Process": f"""This metric evaluates whether the model's responses align with the target's consistency in "Situation understanding (S) - Internal state (I) - Response strategy (B)" under specific contexts. The focus is: can it correctly read the situation, go through reasonable internal states, and adopt response strategies consistent with the persona.
Error types:
1. Situation understanding error: Failure to correctly understand the current situation, such as interpreting support as offense, or neutral as praise or vice versa.
2. Internal state contradiction: The model's inferred internal state doesn't match the target's reaction patterns described in the persona.
3. Response strategy error: Adopting response strategies that don't belong to the target in the persona."""}

\end{Verbatim}
\end{tcolorbox}
\vspace{-1mm}
\caption{Evaluation rubrics used for all datasets}
\label{figure:evaluation_rubrics}
\vspace{-3mm}
\end{figure*}
\begin{figure*}[t!]
\begin{tcolorbox}[
  colback=gray!10,
  colframe=gray!80,
  rounded corners,
  width=\linewidth,
  boxrule=0.5mm,
  left=1mm,right=1mm,top=0mm,bottom=1mm
]
\footnotesize\ttfamily
\begin{Verbatim}[breaklines=true,breakanywhere=true,breaksymbolleft= ]
System prompt:
You are an expert in mimicking the speaking style of specific characters. You need to play the role of {person_name}, responding to questions in the character's tone and behavioral habits.

User prompt:
You will play the role of {person_name}. Based on the information provided below, please respond to the question. Please note the following points:
# Objective
- If the historical memory includes specific language habits, please maintain them.
- Output only the text of your reply; do not add any other explanations or titles.

# Persona and Past Interaction Records
{persona}
# Profile of {person_name}
{profile}

- The three aspects of persona are: Memory, Speaking Style, and SIB_triplet. Memory refers to {person_name}'s personal information and domain knowledge. Speaking Style refers to {person_name}'s word choices and catchphrases. SIB_triplet refers to {person_name}'s Internal state and Behavior strategies when facing specific Situations.

# Question
{question}
\end{Verbatim}
\end{tcolorbox}
\vspace{-1mm}
\caption{Reply generation prompt used in RoleAgentBench. All datasets share the same prefix prompt except for the data input format (question).}
\label{figure:reply_prompt}
\vspace{-3mm}
\end{figure*}

\end{document}